\documentclass[twoside,twocolumn,9pt]{extarticle}
\makeatletter
\@ifundefined{iflatexml}{\newif\iflatexml\latexmlfalse}{}
\makeatother
\usepackage[super,sort&compress,comma]{natbib}
\usepackage[left=1.5cm, right=1.5cm, top=1.785cm, bottom=2.0cm]{geometry}
\usepackage{balance}
\usepackage{mathptmx}
\usepackage{graphicx} 
\usepackage{lastpage}
\usepackage[format=plain,justification=justified,singlelinecheck=false,font={stretch=1.125,small,sf},labelfont=bf,labelsep=space]{caption}
\usepackage{float}
\usepackage{fancyhdr}
\usepackage{fnpos}
\usepackage[english]{babel}
\addto{\captionsenglish}{%
  
}
\usepackage{array}
\usepackage[scaled=0.92]{helvet}
\usepackage{charter}
\usepackage[T1]{fontenc}
\usepackage[dvipsnames]{xcolor}
\usepackage{setspace}
\usepackage[compact]{titlesec}
\iflatexml
\usepackage{url}
\providecommand{\href}[2]{#2}
\else
\usepackage{hyperref}
\fi
\makeatletter
\@ifundefined{NoHyper}{\newenvironment{NoHyper}{}{}}{}
\makeatother

\usepackage[printonlyused]{acronym}
\usepackage{amsfonts}
\usepackage{amsmath}
\usepackage{booktabs}
\usepackage{multirow}
\iflatexml
\usepackage[justification=centering]{subcaption}
\newcommand{\orcidlink}[1]{}
\newcommand{\datasetInner}[2]{\fbox{\begin{minipage}{.25\textwidth}\centering\footnotesize Histogram data\end{minipage}}}
\newcommand{\rhoSqHInner}[2]{\fbox{\begin{minipage}{.24\textwidth}\centering\footnotesize Histogram data\end{minipage}}}
\else
\usepackage{pgfplots}\pgfplotsset{compat=1.18}
\usepackage[justification=centering]{subcaption}
\usepackage{orcidlink}
\newcommand{\datasetInner}[2]{
\begin{tikzpicture}[trim axis left,trim axis right]
    \begin{axis}[
        small,
        xmin=0,
        xmax=0.75,
        ymin=0,
        width=.3\textwidth,
        xlabel=area percentage,
        ylabel=normalised frequency,
        y tick label style={
            /pgf/number format/.cd,
            fixed,
            fixed zerofill,
            precision=2,
            /tikz/.cd
        },
        scaled y ticks = false,
        ]
        \addplot [
            ybar interval,
            fill=blue!50!white,
            draw=blue,
        ] file {#1};
    \end{axis}
    \begin{axis}[
        axis y line*=right,
        axis x line=none,
        small,
        xmin=0,
        xmax=0.75,
        ymin=0,
        width=.3\textwidth,
        ylabel=cumulative frequency,
        ]
        \addplot [
            thick,
            draw=red,
        ] file {#2};
    \end{axis}
\end{tikzpicture}}
\newcommand{\rhoSqHInner}[2]{
\begin{tikzpicture}[trim axis left,trim axis right]
    \begin{axis}[
        /pgf/number format/.cd,
        1000 sep={},
        small,
        width=.29\textwidth,
        ybar,
        enlargelimits=true,
        xmin=0,
        ymin=0,
    ]
    \addplot [
        ybar interval,
        fill=#2!50!white,
        draw=#2,
    ] file {#1};
    \end{axis}
\end{tikzpicture}}
\fi

\definecolor{cream}{RGB}{222,217,201}
\newcommand{\aamfooter}{Author accepted manuscript -- Phys. Chem. Chem. Phys.}

\begin{document}
\raggedbottom

\pagestyle{fancy}
\thispagestyle{plain}
\fancypagestyle{plain}{
\fancyhf{}
\fancyfoot[C]{\footnotesize{\sffamily\aamfooter}}
\fancyfoot[R]{\footnotesize{\sffamily\thepage}}
\renewcommand{\headrulewidth}{0pt}
\renewcommand{\footrulewidth}{0pt}
}

\makeFNbottom
\makeatletter
\renewcommand\LARGE{\@setfontsize\LARGE{15pt}{17}}
\renewcommand\Large{\@setfontsize\Large{12pt}{14}}
\renewcommand\large{\@setfontsize\large{10pt}{12}}
\renewcommand\footnotesize{\@setfontsize\footnotesize{7pt}{10}}
\makeatother

\renewcommand{\thefootnote}{\fnsymbol{footnote}}
\renewcommand\footnoterule{\vspace*{1pt}%
\color{cream}\hrule width 3.5in height 0.4pt \color{black}\vspace*{5pt}} 
\setcounter{secnumdepth}{5}

\makeatletter 
\renewcommand\@biblabel[1]{#1}            
\renewcommand\@makefntext[1]%
{\noindent\makebox[0pt][r]{\@thefnmark\,}#1}
\makeatother 
\renewcommand{\figurename}{\small{Fig.}}
\titleformat{\section}{\sffamily\Large\bfseries}{\thesection}{1em}{}
\titleformat{\subsection}{\normalsize\bfseries}{\thesubsection}{1em}{}
\titleformat{\subsubsection}{\bfseries}{\thesubsubsection}{1em}{}
\setstretch{1.125}
\setlength{\skip\footins}{0.8cm}
\setlength{\footnotesep}{0.25cm}
\setlength{\jot}{10pt}
\titlespacing*{\section}{0pt}{4pt}{4pt}
\titlespacing*{\subsection}{0pt}{15pt}{1pt}

\fancyfoot{}
\fancyfoot[RO,LE]{\footnotesize{\sffamily\thepage}}
\fancyfoot[CO,CE]{\footnotesize{\sffamily\aamfooter}}
\fancyhead{}
\renewcommand{\headrulewidth}{0pt} 
\renewcommand{\footrulewidth}{0pt}
\setlength{\arrayrulewidth}{1pt}
\setlength{\columnsep}{6.5mm}
\setlength\bibsep{1pt}

\twocolumn[
  \begin{@twocolumnfalse}
\begin{center}
{\LARGE\bfseries Accurate identification and measurement of the precipitate area by two-stage deep neural networks in novel chromium-based alloys\par}
\vspace{0.8em}
{\large
Zeyu Xia \orcidlink{0000-0003-0234-5857}$^\ddag$\textit{$^{a}$}, 
Kan Ma \orcidlink{0000-0001-5729-5477}$^\ddag$\textit{$^{c}$}, 
Sibo Cheng \orcidlink{0000-0002-8707-2589}$^{\ast}$\textit{$^{b}$}, 
Thomas Blackburn \orcidlink{0000-0001-9160-3285}\textit{$^{c}$},
Ziling Peng \textit{$^{d}$}, 
Kewei Zhu \orcidlink{0009-0007-5858-6533}\textit{$^{e}$},
Weihang Zhang \orcidlink{0000-0002-6244-5748}\textit{$^{b}$}, 
Dunhui Xiao \orcidlink{0000-0003-2461-523X}\textit{$^{f}$},
Alexander J Knowles \orcidlink{0000-0002-3918-4367}\textit{$^{c}$} 
and Rossella Arcucci \orcidlink{0000-0002-9471-0585}\textit{$^{g}$}
}

\vspace{0.8em}
{\small\textit{Author accepted manuscript. Published in \textit{Physical Chemistry Chemical Physics}, 2023, \textbf{25}, 15970--15987. DOI: \href{https://doi.org/10.1039/D3CP00402C}{10.1039/D3CP00402C}.}\par}
{\footnotesize\textit{The published PCCP version is distributed under the Creative Commons Attribution-NonCommercial 3.0 Unported License (\href{https://creativecommons.org/licenses/by-nc/3.0/}{CC BY-NC 3.0}).}\par}
\end{center}
\vspace{0.8em}
\noindent\textbf{Abstract}\quad
The performance of advanced materials for extreme environments is underpinned by their microstructure, such as the size and distribution of nano- to micro-sized reinforcing phase(s). Chromium-based superalloys are a recently proposed alternative to conventional face-centred-cubic superalloys for high-temperature applications, \textit{e.g.}, Concentrated Solar Power. Their development requires the determination of precipitate volume fraction and size distribution using \acf*{EM}, as these properties are crucial for the thermal stability and mechanical properties of chromium superalloys. Traditional approaches to \acs*{EM} image processing utilise filtering with a fixed contrast threshold, which leads to weak robustness to background noise and poor generalisability to different materials. It also requires an enormous amount of time for manual object measurements on large datasets. Efficient and accurate object detection and segmentation are therefore highly desired to accelerate the development of novel materials like chromium-based superalloys. To address these bottlenecks, based on \acs*{YOLO}v5 and SegFormer structures, this study proposes an end-to-end, two-stage deep learning scheme, DT-SegNet, to perform object detection and segmentation for \acs*{EM} images. The proposed approach can thus benefit from the training efficiency of \acl*{CNN}s at the detection stage (\textit{i.e.}, a small number of training images required) and the accuracy of the \acl*{ViT} at the segmentation stage. Extensive numerical experiments demonstrate that the proposed DT-SegNet significantly outperforms the state-of-the-art segmentation tools offered by Weka and ilastik regarding a large number of metrics, including accuracy, precision, recall and F1-score. This model forms a useful tool to aid microstructure examinations in alloy development, and offers significant advantages to address the large datasets associated with high-throughput alloy development approaches.\par

\end{@twocolumnfalse} \vspace{0.6cm}

]

\renewcommand*\rmdefault{bch}\normalfont\upshape
\rmfamily

\begin{NoHyper}
\footnotetext{\textit{$^{a}$~Queensland University of Technology, Queensland 4000, Australia.}}
\footnotetext{\textit{$^{b}$~Data Science Institute, Department of Computing, Imperial College London, London SW7 2AZ, United Kingdom.}}
\footnotetext{\textit{$^{c}$~School of Metallurgy and Materials, University of Birmingham, Birmingham B15 2SQ, United Kingdom.}}
\footnotetext{\textit{$^{d}$~Institute of Advanced Science Facilities, Shenzhen 518107, P. R. China.}}
\footnotetext{\textit{$^{e}$~Department of Computer Science, University of York, York Y010 5DD, United Kingdom.}}
\footnotetext{\textit{$^{f}$~School of Mathematical Sciences, Tongji University, Shanghai 200092, P. R. China.}}
\footnotetext{\textit{$^{g}$~Department of Earth Science and Engineering, Imperial College London, London SW7 2BP, United Kingdom.}}

\footnotetext{$^\ddag$~These authors contributed equally to this work.}
\footnotetext{\textit{$^\ast$~Corresponding author: \href{mailto:sibo.cheng@imperial.ac.uk}{sibo.cheng@imperial.ac.uk}}}
\end{NoHyper}

\section{Introduction}
\label{sec:introduction}
The integration of microstructural and chemical characterization, property evaluation, and numerical tools is essential in modern-day metallurgy to enhance the design, development, and deployment of alloys. This integration is facilitated using the Integrated Computational Materials Engineering \ac{ICME} frameworks and the Materials Genome Initiative \ac{MGI}~\cite{bergIlastikInteractiveMachine2019}. In computational materials science, learning-based approaches have been incorporated into the \ac{CALPHAD} models to enable the high-throughput calculations for \textit{ab initio} modelling, phase boundary identification, and kinetics modelling~\cite{curtaroloHighthroughputHighwayComputational2013, huangMachinelearningPhasePrediction2019,geDeepLearningAnalysis2020}. These approaches not only accelerate the material design in an ``infinite'' material design space, but are also highly desirable to be paired with high-throughput experimental investigations and subsequent data processing for the analysis of novel materials, including their microstructure recognition on large micrograph image datasets.

In the microstructure of many engineering alloys and novel alloys, secondary phases are known to be influential on mechanical behaviour. The volume fraction, size and shape of secondary phases or particles in alloys are, therefore, important parameters. Equipped with an optical microscope or, more frequently today, an \ac{EM}, images of microstructure can be easily acquired, and image-driven microstructure analysis is an essential step to obtain the information of second phases or particles. Accurate segmentation is thus of the utmost importance for microstructure recognition. The most used microstructure segmentation method in material science is the manual selection of thresholds, such as using the most popular free software ImageJ~\cite{hartigBasicImageAnalysis2013}, or using an automatic global thresholding algorithm~\cite{lieversEvaluationGlobalThresholding2004}, but it is not suitable for many cases, especially subtle thresholds for multi-modal histogram images, in other words, images with varying background contrast such as \ac{TEM} images mentioned in Verguet \textit{et al.}'s work~\cite{amandineImageJToolSimplified2019}. Although many computer vision segmentation techniques such as edge detection, region-based segmentation, partial differential equation, and watershed segmentation can improve the accuracy by using more carefully engineered features~\cite{sarmaComparativeStudyNew2021}, they all present limitations such as sensitivity to noise and impractical use for a large amount of data.

Today, machine-learning-based segmentation techniques have been widely applied not only to cell tracking~\cite{ershovTrackMateIntegratingStateoftheArt2022}, brain tumour segmentation~\cite{nishaReviewBrainTumor2018}, autonomous driving~\cite{luSegmentationBasedMultitaskLearning2022, zhouContextAwareMixupDomain2022}, and geographic segmentation~\cite{wangCBAMBasedMultiscale2022,Cheng2022JCP,cheng2022parameter}, but also to material science~\cite{holmOverviewComputerVision2020}. DeCost et al.~\cite{decostComputerVisionApproach2015} adopted the “bag of visual features” image representation for an \ac{SVM} model to perform microstructure classification. Based on \ac{FCNN}, Azimi et al.~\cite{azimiAdvancedSteelMicrostructural2018} proposed a robust method to classify certain microstructural constituents of low carbon steel for steel quality appreciation. DeCost et al.~\cite{decostHighThroughputQuantitative2019} proposed a DCNN-based model to perform segmentation on complex microstructures. Ma et al.~\cite{maDeepLearningBasedImage2018} proposed a local processing method and a symmetric rectification so that their base model, DeepLab, outperforms existing segmentation models. Inspired by U-Net, Roberts et al.~\cite{robertsDeepLearningSemantic2019} proposed the CNN-based DefectSegNet to perform crystallographic defects segmentation in structural alloys. Cohn et al.~\cite{cohnInstanceSegmentationDirect2021} proposed an instance segmentation tool for metal powder particles produced from gas atomization based on Mask-RCNN, so that researchers can measure the distribution of particle sizes, as well as measure the satellite content in powder samples. Recently, the segmentation for precipitate analysis using the machine learning tool has been attracting increasing attention. Liu et al.~\cite{liuEvolutionAnalysisPrecipitate2022} proposed a CNN-based model to identify materials descriptors describing $\gamma^\prime$ precipitate coarsening in Co-based superalloys. Wang et al.~\cite{wangLearningPrecipitatesMorphological2021} adopted the U-Net segmentation model and a regression model to predict the morphological parameters of the microstructure. Wang et al.~\cite{wangDeepLearningbasedApproach2022} proposed a framework that consists of a U-Net module and ResNet50 module to detect $\delta$ phase and estimate its area accurately. Software packages integrated with common segmentation models like ilastik pixel classification~\cite{sommerIlastikInteractiveLearning2011,bergIlastikInteractiveMachine2019} and Weka trainable segmentation~\cite{arganda-carrerasTrainableWekaSegmentation2017} have achieved microscopy pixel classification tasks in material science. This emerging topic is attracting increasing attention, and it holds promise for precipitate analysis.
Although previous models yielded successful segmentation results, the algorithms used in these models were not state-of-the-art. We propose the implementation of state-of-the-art models like the \ac{YOLO} detection model and SegFormer segmentation model, which will allow for higher efficiency and accuracy in segmentation. Efficient and accurate measurement of precipitate size is imperative for the analysis of precipitate size evolution during the ageing heat treatment, which determines their coarsening rate. In addition, the comparison between the previous models and models to date for precipitate analysis has not been addressed.

Given that precipitates have, in general, a regular shape, \textit{e.g.} spherical or cuboidal, a general dataset containing different conditions of microstructures can be created from existing samples of materials to train a deep learning model, which can then intelligently perform the analysis in new datasets. In this context, this work highlights the application of a deep learning method to precipitate detection in the microstructural design of materials for high-temperature applications. High-temperature materials, including \ac{fcc} nickel-based and cobalt-based superalloys, undergo precipitation during heat treatment, leading to precipitate strengthening\cite{reedSuperalloysFundamentalsApplications2008, callisterFundamentalsMaterialsScience2000}. In these state-of-the-art materials, the precipitate volume fraction and size distribution after different heat treatments are crucial for the strength and creep resistance of such alloys. The coarsening of precipitates in \ac{fcc}-superalloys has been extensively studied~\cite{gesCoarseningBehaviourNibase2007,zhaoGammaPrimeCoarsening2004,meherCoarseningKineticsPrecipitates2013,sauzaMicrostructuralEvolutionHightemperature2019} and enable the precise control of their microstructure and desired properties. Developing novel materials, such as \ac{bcc} chromium-based~\cite{doganCoherentPrecipitationHightemperature2014,locqQuaternaryChromiumbasedAlloys2015} and iron-based ferritic superalloy~\cite{dorantes-rosalesPrecipitationProcessFeNiAlBased2015,sunNanoSizedPrecipitateStability2015}, also requires extensive microstructural observations after various heat treatments using \ac{EM} and lengthy data processing times. Image processing refers to identifying the matrix and precipitate phases, followed by measuring the size distribution and area fraction of the precipitate.

Cr-superalloys, principally \ac{Cr}--\ac{NiAl} alloys consisting of a disordered \ac{bcc} Cr matrix with an A2 structure strengthened by ordered \ac{bcc} \ac{NiAl} intermetallics with a B2 structure, have been identified as potential alternatives to nickel-based superalloys and advanced austenitic steels for high-temperature applications~\cite{doganMicrostructuralStudyHightemperature2013,doganCoherentPrecipitationHightemperature2014,locqQuaternaryChromiumbasedAlloys2015}. Cr-superalloys with \ac{Fe} additions have been further developed in the framework of a European project COMPASsCO2 for advanced Concentrated Solar Power
applications~\cite{baikEffectHafniumMicroaddition2018}. \ac{Cr} offers advantages such as a high melting point, low cost, good oxidation resistance, and low mass density. However, \ac{Cr}--\ac{NiAl} alloys are a nascent class of materials, and their precipitate coarsening kinetics are yet to be investigated.

The size of the B2 precipitates and their morphology are important for the mechanical behaviour of these NiAl-strengthened alloys, such as achieving a high yield strength or creep resistance~\cite{songFerriticAlloysExtreme2015,sunNewDesignAspects2013} in Fe--NiAl ferritic alloy systems. Studying the coarsening rate also contributes to the evaluation of material parameters of new alloys, such as interfacial energy and diffusion coefficients, which will be utilised in physical models for \ac{CALPHAD} and \ac{ICME}. However, the precipitate coarsening alongside the structure-property relationship is principally unknown for Cr-superalloys. Moreover, calculating coarsening rates requires the measurement of precipitate size in numerous samples aged at various temperatures and ageing times, which is laborious through traditional methods.

In this paper, a new, robust, and accurate 2-stage segmentation model on novel $\beta\text{--}\beta^\prime$ chromium-based alloys (Cr-superalloys for short) is proposed. This work aims to develop a learning-based approach to investigate the precipitate area and size distribution in Cr-superalloys. In summary, this paper aims to highlight the following:

\begin{itemize}
    \item manufacture of Cr-superalloys with various heat treatments to produce an A2-B2 microstructure with B2--NiAl sizes varying from the $\text{nm}$ to $\mu\text{m}$ scale.
    \item development of an end-to-end object segmentation model using a two-stage \ac{DNN} DT-SegNet for object segmentation on \ac{EM} images with separate training of the detection and segmentation networks.
    \item application of the DT-SegNet to determine the area fraction and size distribution of precipitates in Cr-superalloys.
    \item demonstration that the developed DT-SegNet can outperform the state-of-the-art segmentation methods in terms of F1-score.
\end{itemize}

\section{Material and methodology}
\label{sec:material-methodology}

\subsection{Studied materials}
\label{subsec:studied-materials}

\begin{table*}[htb]
    \small
    \caption{\ Cr-superalloy sample compositions in atomic percent ($at.\%$) and their respective heat treatment conditions}
    \label{tbl:samples}
    \begin{tabular*}{\textwidth}{@{\extracolsep{\fill}}lllll}
    \hline
    Label & Composition & $\text{Heat Treatment}^\star$ & Phases Expected & \acs{SEM} Observation \\
    \hline
    5-5 & Cr--5Ni--5Al & H+A1 & A2/B2 & Matrix - precipitates \\
    5-5-10 & Cr--5Ni--5Al--10Fe & H+A2 & A2/B2 & Matrix - precipitates \\
    10-10-20-4h & Cr--10Ni--10Al--20Fe & H+A1 & A2/B2 & Matrix - precipitates \\
    10-10-20-100h & Cr--10Ni--10Al--20Fe & H+A3 & A2/B2 & Matrix - precipitates \\
    \hline
    \end{tabular*}
    \raggedright{\small$^\star\text{Heat treatment annotation}$ \\ H: Homogenisation at $ 1400^\circ \text{C} $ for 20 hours \\ A1: Ageing at $ 1200^\circ \text{C} $ for 4 hours \\ A2: Ageing at $ 1000^\circ \text{C} $ for 100 hours \\ A3: Ageing at $ 1200^\circ \text{C} $ for 100 hours}
\end{table*}

\begin{figure*}[htb]
    \centering
    \includegraphics[width = 0.8\textwidth]{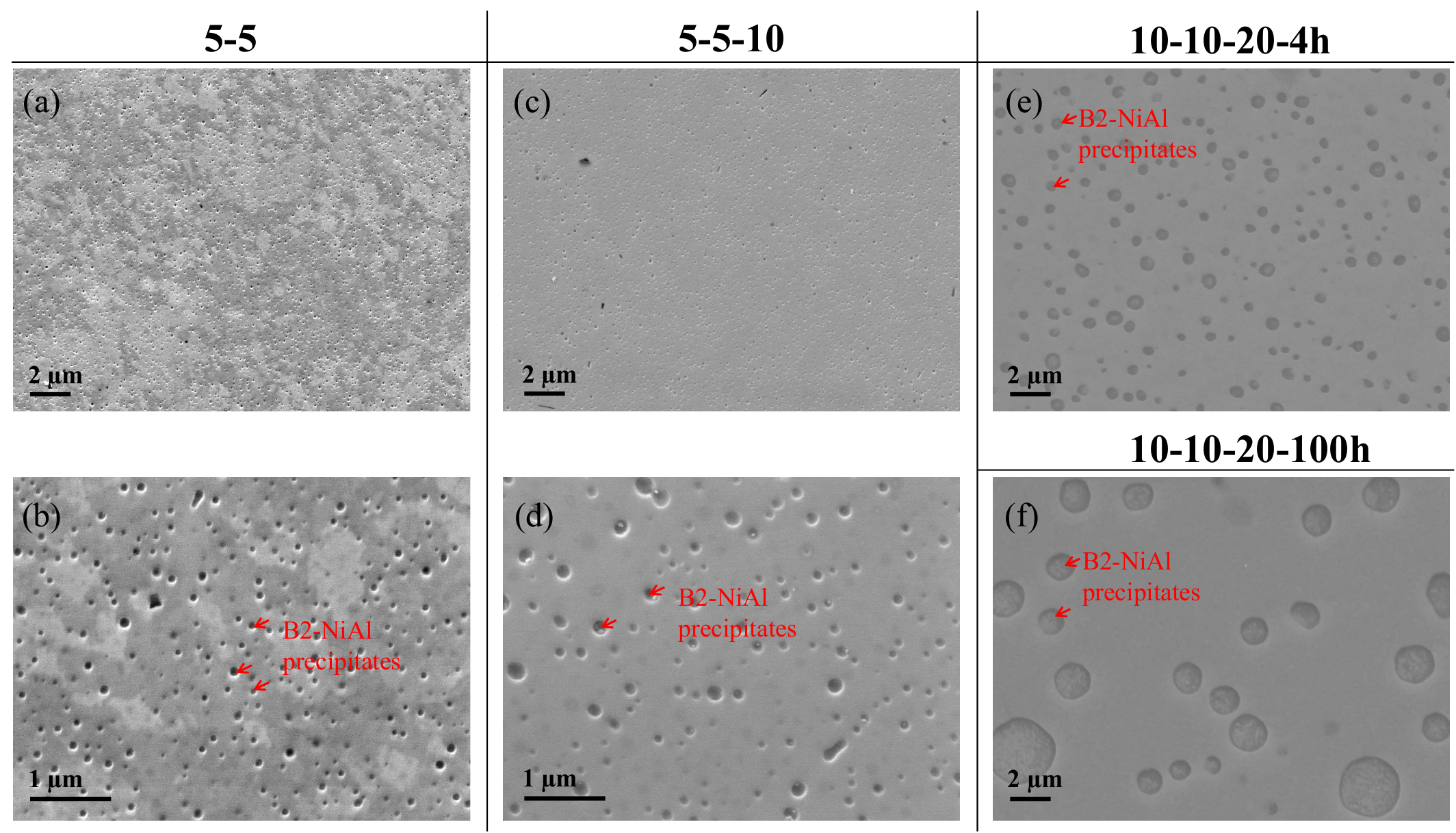}
    \caption{\ac{SEM} micrographs showing the general microstructure of (a) Cr--5Ni--5Al, (c) Cr--5Ni--5Al--10Fe, (e) and (f) Cr--10Ni--10Al--20Fe aged differently. (b) and (d) are zoomed images respectively of (a) and (c).}
    \label{fgr:fig1}
\end{figure*}

\begin{figure*}[htb]
    \centering
    \includegraphics[width = 0.9\textwidth]{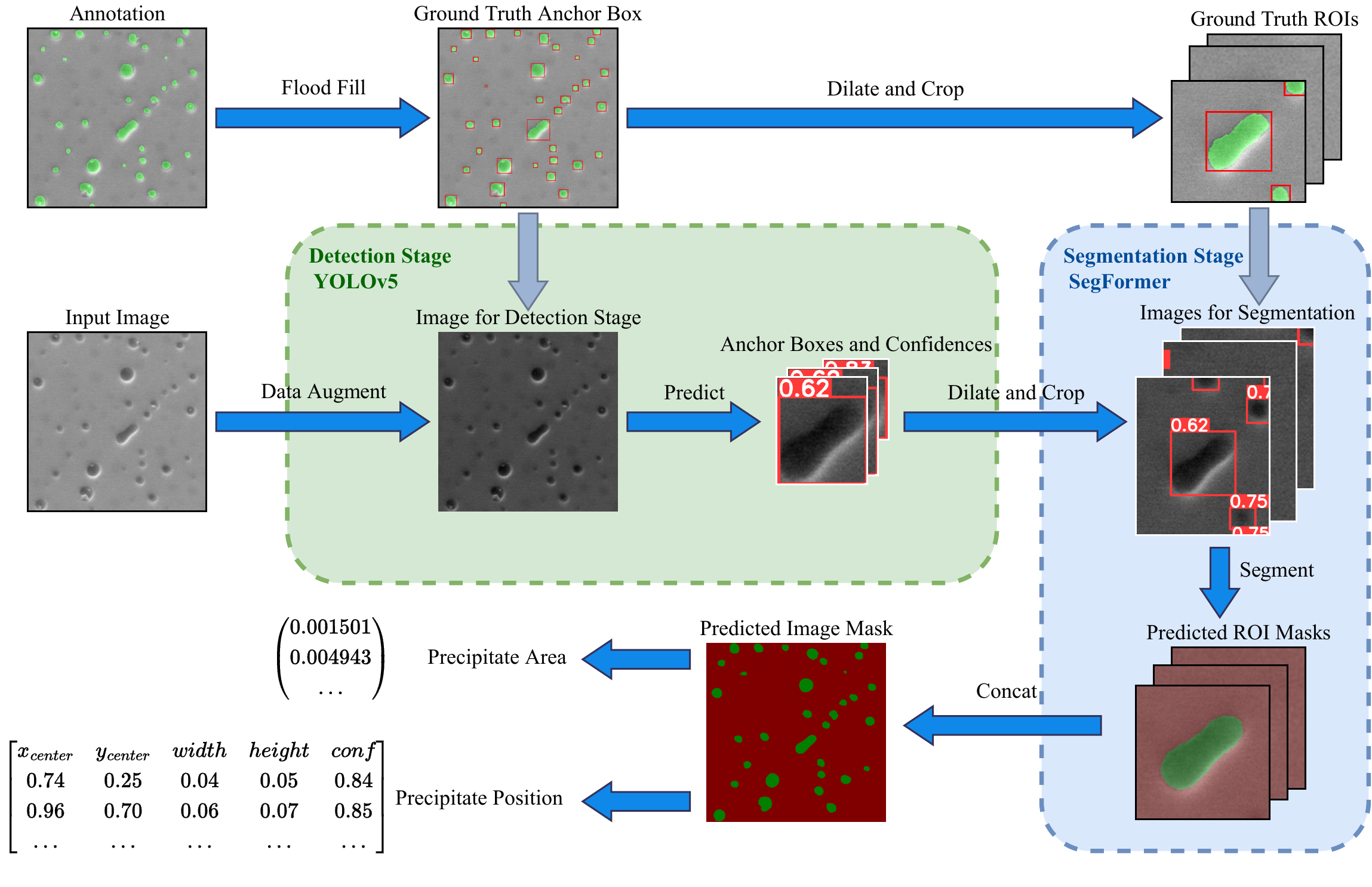}
    \caption{Architecture and pipeline of the proposed DT-SegNet. The first step is to pre-process the input image to fit the input size of the detection network. Once predicted, the anchor boxes are dilated and cropped before feeding into the segmentation network. Finally, the accurate mask, area and position of precipitate objects are derived.}
    \label{fgr:pipeline}
\end{figure*}

After ageing, B2--NiAl spherical precipitates are observed in the \ac{SEM} in all samples, as shown in Fig.~\ref{fgr:fig1}. The size of precipitates varies from nano-scale to micro-scale depending on ageing conditions. The contrast of the precipitates and matrix phases also varies due to the polishing effect on different precipitate sizes. Six \ac{SEM} images, taken at a suitable magnification to contain tens of precipitates, are captured for each sample and used to train the model. In those images, precipitates with their boundaries were carefully identified and manually labelled for the training of the models, as illustrated in Fig.~\ref{fgr:pipeline}. Since most precipitates had a spherical morphology, their sizes were approximately calculated as a function of their radius $r=\sqrt{A/\pi}$ with $A$ being the measured area.

\subsection{The proposed model: DT-SegNet}
\label{subsec:proposed-model}

Driven by the analysis of previous methods, we proposed a novel end-to-end two-stage deep learning scheme combining a Detection (\textbf{DT}) stage and a \textbf{Seg}mentation stage \textbf{Net}work, termed as \textbf{DT-SegNet}. As shown in Fig.~\ref{fgr:pipeline}, the network is designed for precipitate identification and measurement in two stages: a detection stage based on \ac{YOLO}v5~\cite{jocherUltralyticsYOLOv5V62022} and a segmentation stage based on SegFormer~\cite{xieSegFormerSimpleEfficient2021}.

The \ac{YOLO} model is an end-to-end object-detection model which processes the images in the form of small grid regions. Calculating the target bounding boxes and confidences based on weights in smaller regions is crucial to accelerating and enhancing detection accuracy. The SegFormer is a segmentation network consisting of a hierarchical Transformer Encoder backbone, an all-\ac{MLP} decoder neck, and an \ac{MLP} segmentation head. This design allows effective multi-scale extraction and utilisation of critical features without using complex decoders to improve performance and reduce computational costs.

The first detection stage aims to locate the anchor boxes of precipitates with their confidence. In this stage, the input shape of \ac{EM} images is resized to $1280 \text{px} \times1280 \text{px}$. Appropriate data augmentations (such as random scaling, random flipping, mosaic and normalisation) are applied to alleviate the lack of generalisation caused by limited training data. After pre-processing and augmentation, the image is delivered to a \ac{YOLO}v5 network to produce a list of predicted regions with their confidence.

In the second segmentation stage, regions are filtered by a hyper-parameter of the confidence threshold to remove falsely detected regions caused by background noises. To include background information, detected regions are then dilated by 50\% of the original size. Once each extended region is cropped, the new region with extra background information is referred to as the \ac{ROI}, which acts as the input for the SegFormer model. The segmentation model then performs the semantic segmentation task, producing a pixel-wise mask of each precipitate.

Finally, a list of all detected precipitates with their regions, positions and masks can be used to perform precipitate area calculations and other downstream tasks. The overall pipeline is shown in Fig.~\ref{fgr:pipeline}.

\subsubsection{Detection stage.}
\label{subsubsec:detection}

Traditional region proposal neural networks, like Mask R-CNN~\cite{heMaskRCNN2017} and \ac{CNN}~\cite{krizhevskyImageNetClassificationDeep2017}, use bounding boxes and classify detected objects in two stages, resulting in a more extensive computation cost and less awareness of global features. Also, as they scan the whole image with a multi-scale sliding window, the number of windows needs to be pre-defined. Unsatisfactory regions may be detected if only a fixed number of window templates are applied. Compared with two-stage methods, the one-stage YOLO model directly uses joint grid regression to predict both the confidence and the bounding box, which is extremely fast and can learn more generic features of the target object~\cite{zhaoObjectDetectionDeep2019}.

\ac{YOLO} is a family of end-to-end networks for object detection. The \ac{YOLO}v1~\cite{redmonYouOnlyLook2016} is the first end-to-end differentiable neural network which combines object classification and object detection. The author of \ac{YOLO}v3~\cite{redmonYOLOv3IncrementalImprovement2018} added connections to the backbone network layers, which enables the prediction to be made at three different levels of granularity, resulting in a significant performance gain on small objects. \ac{YOLO}v4~\cite{bochkovskiyYOLOv4OptimalSpeed2020} uses new features, including \ac{CSP} connections, cross mini-batch normalisation, self-adversarial-training, mosaic data augmentation and complete \ac{IoU} loss to improve the accuracy and detection speed significantly. \ac{YOLO}v5~\cite{jocherUltralyticsYOLOv5V62022} is the first \ac{YOLO} implementation using the PyTorch framework instead of the Darknet framework. Its novel design includes adaptive anchor boxes, allowing the network to select the most optimal anchor box that fits the dataset. One of the most significant improvements of \ac{YOLO}v5 is its 6x6 Conv2d layer, which reduces the number of parameters without impacting model performance. To increase the inference speed, it also replaces the \ac{SPP} structure with \ac{SPPF}, which is faster with the same output.

\begin{figure*}[htb]
    \centering
    \includegraphics[width = 0.8\textwidth]{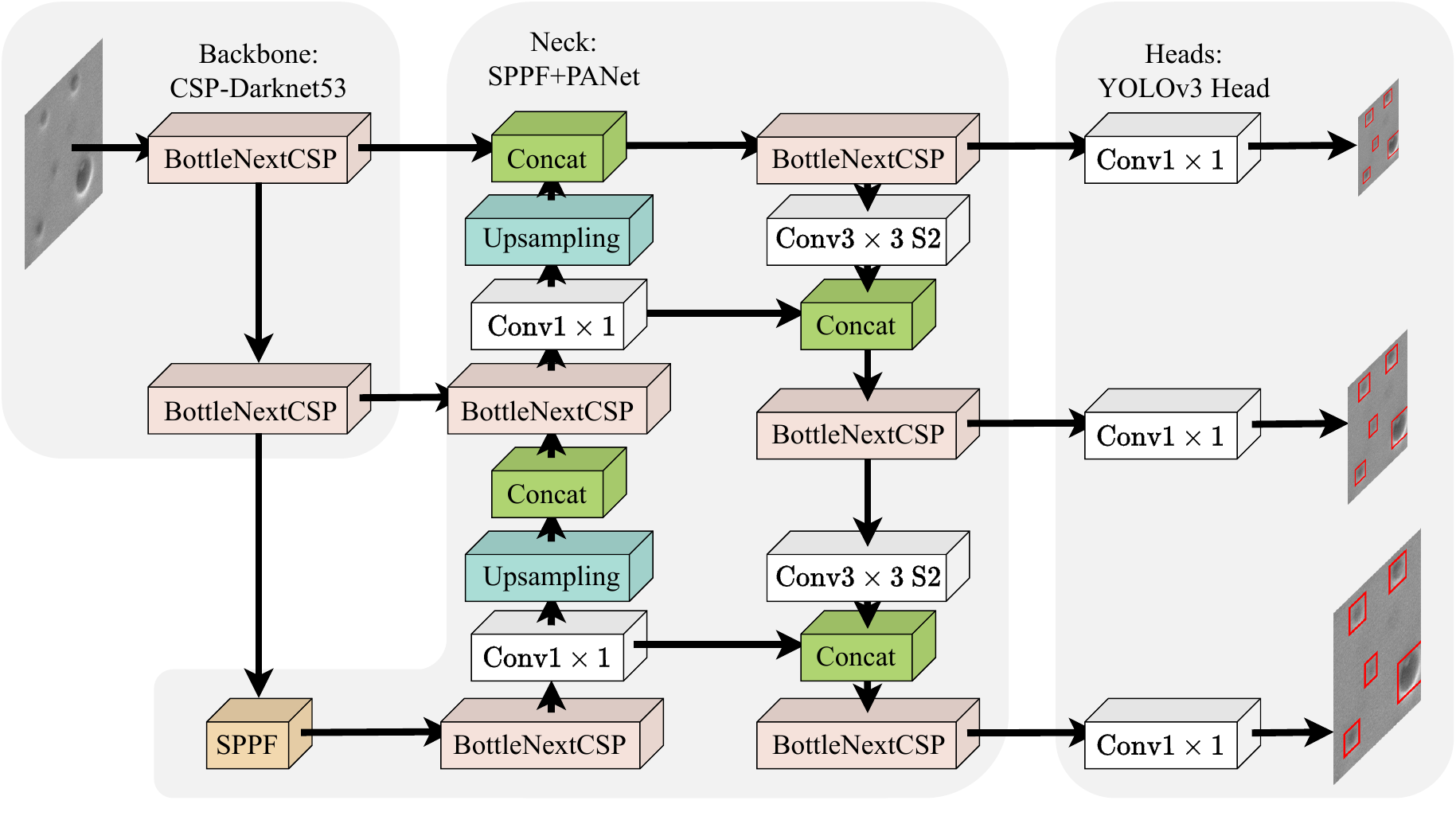}
    \caption{Illustration of the detection stage \ac{YOLO} model. The model consists of three parts: backbone, neck and heads. The backbone extracts features, the neck performs feature fusion, and the heads detect the object.}
    \label{fgr:detsketch}
\end{figure*}

An overview of the \ac{YOLO} model architecture is shown in Fig.~\ref{fgr:detsketch}. \ac{YOLO}v5 is a \ac{CNN}-based one-stage object detection network consisting of a backbone of \ac{CSP}-Darknet53~\cite{wangCSPNetNewBackbone2020}, a neck of \ac{SPPF} and \ac{PANet}~\cite{liuPathAggregationNetwork2018}, and three \ac{YOLO}v3 heads. As seen in the figure, the backbone extracts influential features from input images, and then the neck aggregates all the captured features. Finally, the locations of the objects are computed by the heads. Three heads calculate bounding boxes and probability maps in the grid system and then use all predictions to calculate the final prediction. In summary, \ac{YOLO}v5 adopts all these state-of-the-art techniques in its user-friendly code base, resulting in an outstanding performance with fast speed~\cite{jocherUltralyticsYOLOv5V62022}. Its detection functionality and the ability to detect multi-scale objects benefit our task.

\ac{YOLO}v5 has five models in different scales, all having the same model architecture. The authors designed two parameters: ``depth\_multiple'' and ``width\_multiple'', to control the model scale by multiplying pre-defined constants by the depth and the number of convolutional kernels. This simple design enables selecting the network scale based on the specific problem scale without changing the overall architecture. In this study, multiple networks are tested. After comparing each network, the backbone based on the pre-trained YOLOv5l model with an input size of $1280 \text{px} \times1280 \text{px}$ is selected for the detection stage. A further explanation of the detection model selection is in Section~\ref{subsec:det-discussion}.

The input of the detection stage is a single-channel 2D image. In order to fit all data onto a standard scale, data augmentation is applied to the dataset. The images are resized to $1280 \text{px} \times 1280 \text{px}$ to maintain a consistent network input shape. The output of the detection stage is a list of target anchor boxes for each precipitate. Each anchor box, with corresponding confidence, is represented in the \ac{YOLO} format (x-centre, y-centre, width, height, and confidence).

In this study, improving the detection performance on the small-scale dataset is essential. YOLOv5 utilises several data augmentations to make the most use of the dataset. By applying a set of data augmentations, it is possible to improve the performance without decreasing inference speed~\cite{bochkovskiyYOLOv4OptimalSpeed2020}. In addition to common data augmentation strategies like random scaling, cropping, and random arranging, YOLOv5 introduces two more strategies: Mosaic (first introduced in YOLOv4) and Mixup, which significantly improve the detection accuracy of small objects. Following Bochkovskiy's work~\cite{bochkovskiyYOLOv4OptimalSpeed2020}, four training images are concatenated to allow object detection outside their ordinary context. Batch normalisation~\cite{ioffeBatchNormalizationAccelerating2015} is applied on the concatenated image to reduce the need for a large mini-batch size. This strategy helps generalise the target object by learning the most common features of the target object. Mixup~\cite{zhangMixupEmpiricalRisk2018} is another principle to enhance training performance. By generating convex combinations of different sample images, it regularises the network to select simple linear behaviours to be robust to adversarial inputs. However, since the information of precipitates lies on their edge and internal-external difference, the mixup operation causes a loss of these essential attributes. Therefore, the mixup operation is excluded from our data augmentation method set.

\subsubsection{Segmentation stage.}
\label{subsubsec:segmentation}

\begin{figure*}[htb]
    \centering
    \includegraphics[width = 0.6\textwidth]{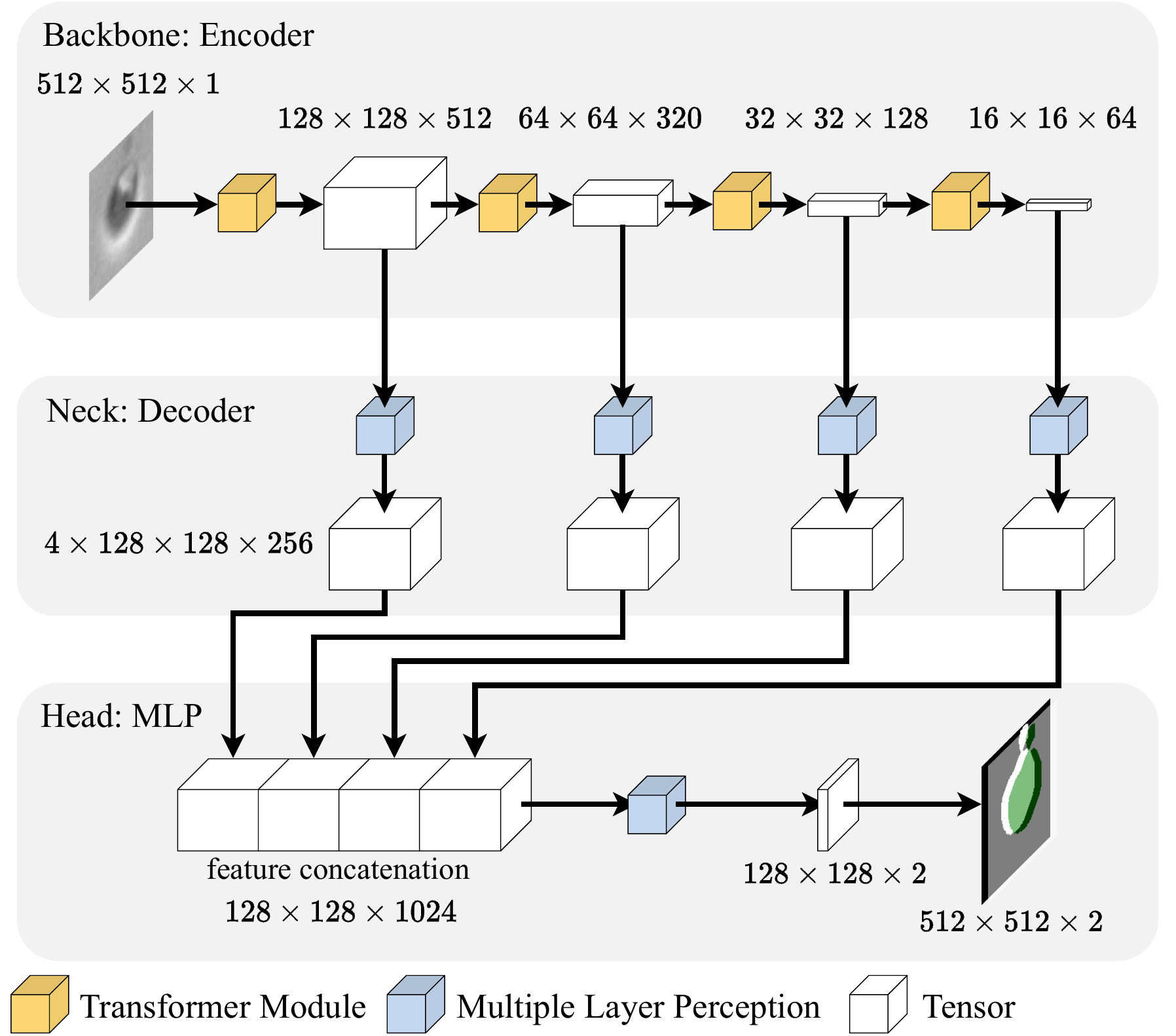}
    \caption{Illustration of the segmentation stage SegFormer model. This model consists of four Transformer modules as an Encoder backbone, an all-\ac{MLP} module as the decoder neck and an \ac{MLP} module as the head.}
    \label{fgr:segsketch}
\end{figure*}

As can be seen from Fig.~\ref{fgr:segsketch}, the network for the segmentation stage, SegFormer~\cite{xieSegFormerSimpleEfficient2021}, consists of an Encoder of four Transformer~\cite{vaswaniAttentionAllYou2017} modules as the backbone, an \ac{MLP} decoder as the neck and an \ac{MLP} segmentation head. Four Transformer modules' backbone extracts coarse-grained and fine-grained features. After that, the neck fuses the extracted features and passes them to the segmentation head so that the head can make a final prediction of the semantic segmentation mask. In the proposed DT-SegNet, essential features such as edges and internal textures are captured and generalised, enabling more precise pixel classification and segmentation on edges with good noise resistance.

Research on \ac{ViT}~\cite{dosovitskiyImageWorth16x162021a} has suggested that a Transformer directly applied to images performs significantly better than traditional \ac{CNN} networks. However, the columnar structure of such a model makes it computationally expensive. Additionally, \ac{ViT} only outputs feature maps of a fixed resolution, which can cause inaccuracy in the segmentation task. To solve these problems, SegFormer~\cite{xieSegFormerSimpleEfficient2021} proposed a simple and efficient design that unifies the Transformer module with lightweight \ac{MLP} decoders. This design achieves excellent performance gains while maintaining a reasonable computation cost.

Although the shape, internal texture, and edge brightness vary between precipitates, most can be detected by their edges. Therefore, fully extracting the edge and perceiving more background information can help distinguish edges from the background. Thus, image dilation is designed ahead of the segmentation stage. In this operation, the boundary of each target anchor box is expanded twice in both width and height, then resized to $ 512\text{px}\times512\text{px}$. The necessary edge information can be kept by applying dilation, making the segmentation stage less sensitive to false precipitate detection. The extra background information also helps the segmentation network to have more information about the context of the target object. The dilated region with extra background information is named \ac{ROI} in this paper.

\section{Dataset}
\label{sec:dataset}

This article conducts experiments on the dataset generated in this study, which contains $ N=24 $ \ac{SEM} two-dimensional images. Details of the dataset are shown in Table~\ref{tbl:dataset}. The data is split into training, validation and test sets (in a 6:2:2 ratio) using the hold-out method to ensure even distribution in each set. Due to the small scale of the dataset, images with similar image features were manually assigned into different sets. By doing this, a comparison of the robustness of different models can be made. The result of evaluating the precipitate areas in both the original image and the \ac{ROI} shows that the dilation operation significantly improves the precipitate area percentage. 

\begin{table}[htb]
    \small
    \caption{\ Statistics of the three datasets split by the ratio of 6:2:2. The precipitate area ratio in ROI was significantly higher than in the raw input image}
    \label{tbl:dataset}
    \begin{tabular*}{0.48\textwidth}{@{\extracolsep{\fill}}lllll}
        \hline
        \multirow{2}{*}{Dataset} & \multirow{2}{*}{Image Count} & \multicolumn{3}{c}{Precipitates} \\
        \cmidrule(lr){3-5}
        & & Count & \% in Image & \% in \acs{ROI} \\
        \hline
        Training & 15 & 1674 & 9.69 & 23.21 \\
        Validation & 4 & 355 & 6.34 & 21.86 \\
        Test & 5 & 243 & 9.73 & 25.64 \\
        \hline
    \end{tabular*}
\end{table}

\begin{figure}[htb]
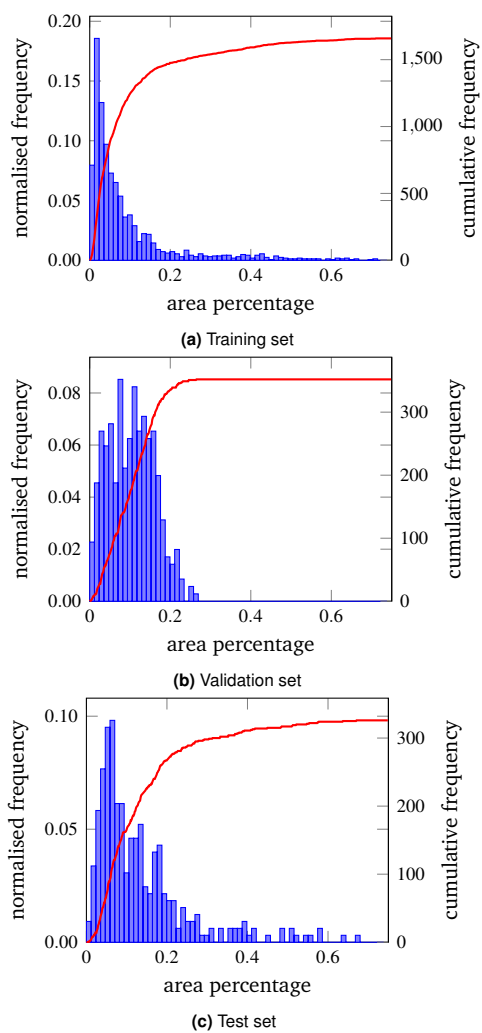

    \centering
    \subcaptionbox{Training set}
    {\datasetInner{hist2_train.txt}{cf_train.txt}\vspace{-.5\baselineskip}\hspace{1.5\baselineskip}}
    \label{fgr:dataset-train}
    \subcaptionbox{Validation set}
    {\datasetInner{hist2_val.txt}{cf_val.txt}\vspace{-.5\baselineskip}\hspace{1.5\baselineskip}}
    \label{fgr:dataset-val}
    \subcaptionbox{Test set}
    {\datasetInner{hist2_test.txt}{cf_test.txt}\vspace{-.5\baselineskip}\hspace{1.5\baselineskip}}
    \label{fgr:dataset-test}
    \caption{Distribution of precipitate scales in the three sets. The bar shows the normalised frequency on each dataset, and the curve shows the cumulative frequency. All three datasets have the most precipitates with areas under 0.2\% of the total area.}
    \label{fgr:dataset}
\end{figure}

The bar charts in Fig.~\ref{fgr:dataset} show the distribution of precipitate scales in three datasets. It can be observed that in all the datasets, most of the precipitate area percentage is under 0.2\%. However, the training set has few aberrant precipitates with relative scales larger than 0.2\%. As for the validation set, the distribution shows a narrower overall range of 0.3\%. The test set contains a set of images where most of the precipitate scales are below 0.2\%, whereas some irregular samples with large scales exist.

A three-phase process is followed to produce ground truth for this dataset. Initially, images are labelled interactively using PaddleSeg~\cite{liuPaddleSegHighefficientDevelopment2021}, and then manually refined using Adobe Photoshop. The shapes and boundaries are corrected during this process. Once finished, the segmentation labels are converted into YOLO-format anchor boxes using the flood-filling algorithm. The final stage comprises a precipitate region correcting step using LabelImg~\cite{tzutalinLabelimg2015}. In this process, overlapping anchor boxes are separated into individual anchor boxes.

\section{Results and discussion}
\label{sec:results-discussion}

To thoroughly evaluate the performance of DT-SegNet, this article first experiments with multiple sets of settings on both the YOLOv5~\cite{jocherUltralyticsYOLOv5V62022} network and the SegFormer~\cite{xieSegFormerSimpleEfficient2021} network to find the most optimised backbone configuration. Then, this article selects five representative methods implemented in two software and four state-of-the-art \ac{CNN} models in the field of general image segmentation as a comparative experiment. Lastly, this article performs a visualisation analysis on four test images to explain the outcome of each method.

\subsection{Implementation details}
\label{subsubsec:implementation}

This model is implemented based on the official PyTorch YOLOv5 v6.1 implementation~\cite{jocherUltralyticsYOLOv5V62022} and PaddleSeg v2.7~\cite{liuPaddleSegHighefficientDevelopment2021} using the PaddlePaddle framework.

At the detection stage, auto-detection of the batch size is used. Minimum epochs of 300 are performed with an early-stopping regularisation of 150-epoch patience. The checkpoint is kept at each epoch. A compound cost function of objectness score, class probability score, and bounding box regression score, a \ac{SGD} optimiser of 0.01 learning rate and a learning rate scheduler of LambdaLR are used. At this stage, data augmentation of mosaic, copy-paste, random scaling, flipping, hue, saturation adjustment, and normalisation processes are used. Due to the limitation in the dataset scale, the official pre-trained model on the \ac{COCO} 2017 dataset~\cite{linMicrosoftCOCOCommon2014} is used for the model to learn more general object features. This dataset includes 80 classes of images with labels such as human, bicycle, traffic light, bird, food, and book.

At the segmentation stage, a batch size of 1, a maximum of 80000 training epochs, and a checkpoint save interval of 200 are used. CrossEntropyLoss cost function, the AdamW optimiser ($\beta_1=0.9\text{, }\beta_2=0.999\text{, weight decay}=0.01$) and a PolynomialDecay learning rate scheduler with learning rate 0.00006 are adopted in our experiments. All images are normalised and applied with random horizontal and vertical flips at this stage. Pretrained MixVisionTransformer models on ImageNet-1K dataset~\cite{russakovskyImageNetLargeScale2015} are used.

Other hyper-parameters from both models are maintained as default in their original implementation. The model with the best performance on the validation set is selected as the best model.

The ``varying contrast'' means the difference between foreground and background pixels varies. Traditional methods that apply a constant threshold or a cross-correlation with a Gaussian window~\cite{gonzalezDigitalImageProcessing2018} provided by the OpenCV library struggle to handle this problem well. In our work, we used normalisation in the data pre-processing pipeline to maximise the margin of different classes of pixels. Then, the encoder module in our network can perform detection and segmentation tasks from images with different contrasts.

\subsection{Baseline approaches}
\label{subsubsec:baseline}

In this study, several widely utilised machine learning methodologies, namely, \ac{FRF} and \ac{MLP} in the Weka software~\cite{arganda-carrerasTrainableWekaSegmentation2017}, and \ac{LDA}, \ac{RF}, and \ac{MLP} in ilastik software~\cite{bergIlastikInteractiveMachine2019}, are deemed as foundational models for comparison purposes. This study also includes a comparative analysis of contemporary state-of-the-art end-to-end deep learning networks, including U-Net, UNet 3+, DeepLabV3+ and SegFormer. The proposed DT-SegNet scheme is compared against these methods using the same training and test datasets.

\ac{RF}~\cite{breimanRandomForests2001} is a decision-tree-based learning method. It works by building an ensemble of decision trees based on input features. During prediction, the model combines the prediction from all trees to make a final prediction, resulting in a better generalisation outcome than a single decision tree. \ac{FRF}~\cite{breimanRandomForests2001} is similar to the standard \ac{RF} algorithm, but with some modifications to accelerate its speed and reduce memory usage. Based on Java and implemented in Trainable Weka Segmentation~\cite{arganda-carrerasTrainableWekaSegmentation2017}, it uses a sub-sampling technique to randomly select a subset of the features and instances for each tree in the forest. It also uses a heuristic algorithm to select the best splitting point at each node, which further improves the model speed. \ac{MLP}~\cite{kubatNeuralNetworksComprehensive1999} is a type of neural network composed of multiple layers of fully-connected artificial neurons. It uses a back-propagation algorithm to adjust the weights of each neuron based on the error between model prediction and ground truth. \ac{LDA}~\cite{hastieElementsStatisticalLearning2009} is a statistical technique that finds a linear combination of input features that maximises the separation between different classes. It models the distribution of input features in each class and uses the between-class variance to the within-class variance ratio to calculate the optimal discriminant space for classifying new image pixels. \ac{SVC}~\cite{plattProbabilisticOutputsSupport1999} is a soft-margin classification algorithm using a regularisation parameter of C to control the balance between maximising the margin and minimising the classification error. U-Net~\cite{ronnebergerUNetConvolutionalNetworks2015} is a widely used \ac{CNN} model initially designed to solve biomedical image segmentation challenges. It consists of a contraction path,  an expansion path, and skip connections that allow the expanding path to use information from the contracting path. This enables it to achieve high accuracy and preserve the original spatial resolution. UNet 3+~\cite{huangUNetFullScaleConnected2020} is an extension of the previous U-Net and its variants. By adding more encoder and decoder layers and introducing dense skip connections and deep supervisions, it has achieved state-of-the-art performance on several medical image segmentation benchmarks. DeepLabV3+~\cite{chenEncoderDecoderAtrousSeparable2018} is a \ac{CNN} model that uses a modified atrous spatial pyramid pooling module to capture contextual information over multiple scales and uses a decoder module to produce pixel-wise predictions. SegFormer~\cite{xieSegFormerSimpleEfficient2021} is a segmentation model that uses a Transformer-based Encoder and a Decoder module with multi-scale feature fusion and progressive upsampling.

Weka trainable segmentation~\cite{arganda-carrerasTrainableWekaSegmentation2017} is a machine-learning tool for microscopy pixel classification. This study evaluates the segmentation models of \ac{FRF} and \ac{MLP} on this software. Weka trainable segmentation version 3.3.2 with Fiji ImageJ 1.53t is used. We use the default set of standard deviation $\sigma$ in the Gaussian filter applied during the image pre-processing step in all Weka experiments, which are $ 1.0, 2.0, 4.0, 8.0, \text{ and } 16.00 $. Gaussian blur (5 convolutions with 5 variations of $\sigma$), Sobel filter, Hessian, the difference between Gaussians (combination of all $\sigma$), and membrane projections (kernel size of $ 19 \times 19 $) are selected as classification features. In this experiment, the \ac{FRF} parameter of unlimited max depth, two-decimal-place precision for model output, and two attributes in the random selection is used to generate 200 trees. In this study, the \ac{MLP} parameter settings of a batch size of 10000, disabled decay, a learning rate of 0.3, momentum of 0.2, two decimal places, and a validation stage set the size of 20 with a threshold of 20. Both methods are trained with balance classes enabled, which filters more populated foreground pixel samples and duplicates less numerous background pixel samples.

Ilastik pixel classification~\cite{bergIlastikInteractiveMachine2019} is an interactive machine-learning tool for bio-image analysis. Segmentation models \ac{LDA}, \ac{RF} and \ac{SVC} are evaluated for comparison. In this study, ilastik version 1.4.0rc6 is used.  As ilastik does not provide an interface to tune parameters, all parameters are set as the default value. In the scikit-learn implementation, the default margin parameter C for \ac{SVC} is 1.0, with an RBF kernel and probability estimates enabled. It trains features of Color and Intensity (Gaussian Smoothing), Edge (Laplacian of Gaussian, Gaussian Gradient Magnitude, and Difference of Gaussians), and Texture (Structure Tensor Eigenvalues and Hessian of Gaussian Eigenvalues) for all images using a $\sigma$ of $0.30, 0.70, 1.00, 1.60, 3.50, 5.00 \text{ and } 10.00$. All the methods are implemented on the scikit-learn backend.

Four single-stage segmentation models are trained and inferred using PaddleSeg v2.7~\cite{liuPaddleSegHighefficientDevelopment2021} on the PaddlePaddle framework, with a checkpoint save interval of 100. U-Net is trained with a batch size of 4, a maximum of 40000 training epochs, no pre-trained model and deconvolution disabled. UNet 3+ is trained with a batch size of 2, a maximum of 40000 training epochs, no pre-trained model, batch normalisation enabled, classification-guided module disabled, and deep supervision disabled. DeepLabV3+ is trained with a batch size of 2, a maximum of 80000 training epochs, ImageNet-1K~\cite{russakovskyImageNetLargeScale2015} pre-trained ResNet50\_vd backbone, a dilation rate of (1, 12, 24, 36), and no pre-trained model. SegFormer B0 and B1 are trained with a batch size of 1 and a maximum of 80000 training epochs. CrossEntropyLoss cost function, the AdamW optimiser ($\beta_1=0.9\text{, }\beta_2=0.999\text{, weight decay}=0.01$) and a PolynomialDecay learning rate scheduler with learning rate 0.00006 are adopted in the experiments for SegFormer. All other models except SegFormer are trained with CrossEntropyLoss cost function, a stochastic gradient descent optimiser ($momentum=0.9, weight decay=0.00004$) and a PolynomialDecay learning rate scheduler with learning rate=0.01, end\_lr=0 and power=0.9.

\subsection{Training environment}
\label{subsubsec:training}

All models are trained and inferred on a server with AMD EPYC 7543 CPU, an NVIDIA RTX A5000 graphics card and 32 GB Memory. Experiments are under Ubuntu 20.04 operation system, with the programming language Python 3.8, GPU acceleration kit CUDA 11.6, machine learning framework PyTorch 1.13.1 and PaddlePaddle 2.4. Two baseline methods, Weka and ilastik, are trained on a desktop machine running on Windows 10 version 22H2 with an Intel Core i5-9600KF CPU, NVIDIA Geforce GTX 1080 GPU and 32 GB Memory. Due to the online training nature of Weka trainable segmentation and ilastik pixel classification, directly using pixel-wise annotation exhausts system resources and results in the system not responding. Two discrete reasons emerge from this. First, the software generates computationally-heavy features on extensive pixels at their pre-processing stage. Second, there is limited support for GPU acceleration. Therefore, all images in this dataset are relabelled using built-in tools inside both software packages to solve this problem. As this action may result in a drop in labelling accuracy, the relabelling is repeated twice until all precipitates in the training set are segmented correctly. Another aspect worth noticing is the size of the output model. The trained model of DT-SegNet has a size of ~198MB, compared with ~257MB of the \ac{LDA} model, ~256MB of the \ac{RF} model, and 359MB of the \ac{SVC} model. However, due to the default unlimited max depth, the \ac{FRF} model has a size of 1.19GB. This can make it challenging to deploy such a big model on machines with less memory and CPU power.

\subsection{Metrics}
\label{subsubsec:metrics}
In this study, a robust comparison of the proposed DT-SegNet against the state-of-the-art tools Weka and ilastik is performed using a wide range of detection and segmentation metrics. Manually labelled data are used as ground truth. The algorithm performances of both detection and segmentation stages are evaluated on the test dataset. Precision, recall, and \ac{mAP} are measured for the detection stage. $TP=\text{True positive}$, $TN=\text{True negative}$, $FP=\text{False positive}$, and $FN=\text{False negative}$ are denoted.

In the detection stage, two bounding boxes: the prediction box $P$ and the ground truth box $T$ are first defined. Then \ac{IoU} can be defined as:

\begin{equation}
    IoU=\frac{|{P}\bigcap{T}|}{|{P}\bigcup{T}|}.
\end{equation}

Based on the IoU, the predicted bounding boxes from the detection model can be classified as $TP$ if the \ac{IoU} exceeds the IoU threshold (0.6 as default).

Precision is a metric that measures how accurate the prediction is. It is calculated as follows:

\begin{equation}
    \text{Precision}=\frac{TP}{TP+FP}.
\end{equation}

Recall demonstrates the ability to find all precipitates, \textit{i.e.},

\begin{equation}
    \text{Recall}=\frac{TP}{TP+FN}.
\end{equation}

Since precision or recall alone can not fully characterise the prediction effect of the model, a metric that measures the precision and recall jointly is needed. \ac{AP}~\cite{everinghamPASCALVisualObject2011} is defined as the area under the \ac{PRC}. The formula is defined as follows:

\begin{equation}
    \ac{AP}=\int_{0}^{1} p(r) \,dr
\end{equation}

where $r$ denotes the recall and $p(r)$ denotes the precision in the function of $r$.

However, the result of \ac{AP} is heavily affected by the selection of the IoU threshold. The \ac{mAP} metric~\cite{linMicrosoftCOCOCommon2014} is used to alleviate this problem. This metric calculates the average \ac{AP} score on different \ac{IoU} thresholds. In this task, $\ac{mAP}_{0.5}$ is the \ac{AP} with the \ac{IoU} threshold of 0.5. $\ac{mAP}_{0.5:0.95}$ computes average \ac{AP} using \ac{IoU} thresholds of $[0.5, 0.55, 0.60,\cdots, 0.95]$. Since $\ac{mAP}_{0.5:0.95}$ reflects the model performance under most of the \ac{IoU} thresholds, it is used as the primary metric in the detection stage of this study.

Accuracy, precision, recall, \ac{IoU}, \ac{SSIM}, and F1-score are evaluated in the segmentation stage. At this stage, the $TP$ predictions as pixels predicted are defined to have the same label as the ground truth annotation.

The pixel-wise accuracy for the segmentation stage is defined as:

\begin{equation}
    \text{Pixel accuracy}=\frac{TN+TP}{TN+FP+TP+FN}
\end{equation}

This metric represents the number of correctly segmented pixels over the total number of pixels. The area accuracy is also computed, which is defined as follows:

\begin{equation}
    \text{Area accuracy}=\frac{\text{total precipitate area in prediction}}{\text{total precipitate area in ground truth}}
\end{equation}

This metric conveys the difference between the predicted and actual area. Precision and recall have the exact definition in the detection stage, but the calculation is performed pixel-wise. The mean \ac{IoU} is the average \ac{IoU} on precipitate and background class.

The following formula is used to calculate \ac{IoU} in the segmentation stage:

\begin{equation}
    \ac{IoU}=\frac{TP}{TP+FP+FN}
\end{equation}

\ac{SSIM}~\cite{wangMeanSquaredError2009} is used to measure the similarity between the prediction and the ground truth of the exact shape of the precipitate.

The F1-score, defined as

\begin{equation}
\text{F1}=\frac{2\times{TP}}{2\times{TP}+FP+FN},
\end{equation}

can evaluate both precision and recall. Thus, it is selected as the primary metric for comparing model performances in the segmentation stage.

In summary, precision and recall are general metrics for both the detection and segmentation stages. The \ac{mAP} is used for the detection stage only. Accuracy, \ac{IoU}, \ac{SSIM} and F1-score are used for the segmentation stage.

\subsection{Detection backbone}
\label{subsec:det-discussion}

Multiple combinations of models with two input shapes are experimented to find the best configuration in the detection stage. The effectiveness of transfer learning is also explored by using models pretrained at \ac{COCO} dataset~\cite{linMicrosoftCOCOCommon2014}. All models are trained with patience of 150 epochs. Table~\ref{tbl:detection_performance} shows the performance of different networks with different configurations on the test set. According to Ultralytics, the \ac{COCO} trains natively on $640\text{px}$, and benefits can be obtained from increasing the input image size if a large number of small objects exist in the dataset. The results show that increasing input image size from $640\text{px}$ to $1280\text{px}$ without pre-training may slightly reduce the model performance on small models such as YOLOv5n but increase the performance gain on large models such as YOLOv5s, YOLOv5m and YOLOv5l. And with pre-training, a faster convergence speed with higher performance is discovered, and models perform better in most settings. According to Luo et al.'s work~\cite{luoUnderstandingEffectiveReceptive2016}, the effective receptive field increases when more convolutional layers are added, more pooling layers are placed, or convolution stride is higher. In our cases, \ac{YOLO} networks with an input size of $1280\text{px}\times{1280}\text{px}$ have more convolutional layers than networks with input size $640\text{px}\times{640}\text{px}$. The increased parameters can increase the effective receptive field, thus providing large models with a better generalisation ability on high-resolution input images. The utilisation of pre-trained models shows performance improvement with a 0.6\% increase in $\ac{mAP}_{0.5:0.95}$ on average. The initial weights in the pretrained model may accelerate the gradient descent process in the right direction, thus providing better generalisation ability. After comparison, pre-trained YOLOv5l with an input size of $1280\text{px}\times{1280}\text{px}$ is selected for the detection stage.

\begin{figure}[htb]
    \centering
    \iflatexml
    \fbox{\begin{minipage}{0.4\textwidth}\centering\footnotesize F1-confidence curve data\end{minipage}}
    \else
    \begin{tikzpicture}[trim axis left,trim axis right]
        \begin{axis}[
            width=0.4\textwidth,
            title=F1-Confidence Curve,
            xlabel={Confidence},
            ylabel={F1-score},
            minor x tick num=2,
            minor y tick num=2,
            no markers,
            every axis plot/.append style={ultra thick}
            ]
            \addplot+ [
            ] file {F1_curve.txt};
        \end{axis}
    \end{tikzpicture}
    \fi
    \caption{F1-Confidence Curve of pre-trained YOLOv5l with an input size of $1280\text{px}\times{1280}\text{px}$ on the validation set. Each point on the line indicates the F1-score at the given confidence filter constant. The peak of the F1-score is 0.97, reached at the confidence of 0.475.}
    \label{fgr:f1-curve}
\end{figure}
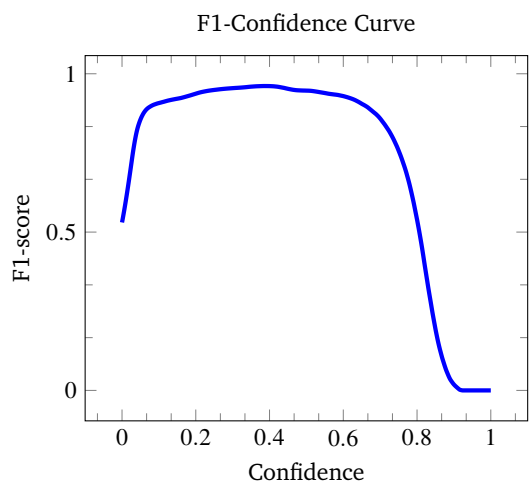

The F1-Confidence curve of YOLOv5 is shown in Fig.~\ref{fgr:f1-curve}. A higher F1-score indicates better detection performance. As seen from the figure, the F1-score reaches its peak at 0.97 with a confidence of 0.475. Furthermore, a wide range of confidence thresholds from 0.1 to 0.6 can be selected to perform precipitate detection.

\begin{table*}[htb]
    \small
    \caption{Detection backbone performance on the test set with different settings}
    \label{tbl:detection_performance}
    \begin{tabular*}{\textwidth}{@{\extracolsep{\fill}}lllllllll}
        \hline
        Backbone & Pre-trained & Batch Size & Epoch & Input Size & Precision & Recall & $\ac{mAP}_{0.5: 0.95}$ & $\ac{mAP}_{0.5}$ \\
        \hline
        YOLOv5n &  & 201 & 494 & $640\times640$ & 96.0 & 94.2 & 58.6 & 97.3 \\
        YOLOv5s &  & 110 & 510 & $640\times640$ & 95.6 & 93.9 & 57.4 & 96.6 \\
        YOLOv5m &  & 64 & 330 & $640\times640$ & 95.6 & 94.3 & 57.0 & 97.0 \\
        YOLOv5l &  & 37 & 645 & $640\times640$ & 97.4 & 94.8 & 57.4 & 97.5 \\
        \midrule
        YOLOv5n &  & 50 & 370 & $1280\times1280$ & 94.7 & 92.9 & 56.0 & 96.2 \\
        YOLOv5s &  & 29 & 426 & $1280\times1280$ & \textbf{97.5} & 93.3 & 59.8 & 97.4 \\
        YOLOv5m &  & 15 & 486 & $1280\times1280$ & 95.2 & 92.0 & 58.1 & 96.9 \\
        YOLOv5l &  & 9 & 424 & $1280\times1280$ & 94.0 & \textbf{96.3} & 61.0 & 98.2 \\
        \midrule
        YOLOv5n & \checkmark & 50 & 366 & $1280\times1280$ & 95.4 & 95.9 & 61.5 & 98.1 \\
        YOLOv5s & \checkmark & 29 & 614 & $1280\times1280$ & 95.9 & 94.5 & 60.6 & 97.7 \\
        YOLOv5m & \checkmark & 15 & 230 & $1280\times1280$ & 91.6 & 93.9 & 52.9 & 97.5 \\
        YOLOv5l & \checkmark & 4 & 400 & $1280\times1280$ & 97.4 & 95.7 & \textbf{62.5} & \textbf{99.0} \\
        \hline
    \end{tabular*}
\end{table*}

\subsection{Segmentation backbone}
\label{subsec:seg-discussion}

For model selection in the segmentation stage, SegFormer B0 and SegFormer B1 are tested with an input image size of $512\text{px}$. Table~\ref{tbl:segmentation_performance} shows the performance of different SegFormer networks on the test set. The result shows that excellent performance is achieved using both models. There is a slight improvement regarding the F1-score in SegFormer B1, which may be attributed to additional parameters in the Encoder. Because both networks achieve outstanding performance on the task, and the computation cost on SegFormer B1 is affordable, SegFormer B1 is chosen as the model in the segmentation stage.

\begin{table*}[htb]
    \small
    \caption{ Segmentation backbone performance on the test set with different scales. P: Precipitate, B: Background}
    \label{tbl:segmentation_performance}
    \begin{tabular*}{\textwidth}{@{\extracolsep{\fill}}llllllllllllll}
        \hline
        \multirow{2}{*}{Backbone} &
        \multirow{2}{*}{Epoch} &
          \multicolumn{2}{c}{Accuracy} &
          \multicolumn{2}{c}{Precision} &
          \multicolumn{2}{c}{Recall} &
          \multicolumn{3}{c}{\acs{IoU}} &
          \multirow{2}{*}{\acs{SSIM}} &
          \multirow{2}{*}{F1} \\\cmidrule(lr){3-4}\cmidrule(lr){5-6}\cmidrule(lr){7-8}\cmidrule(lr){9-11}
         &
         &
          Pixel &
          Area &
          P &
          B &
          P &
          B &
          P &
          B &
          Mean &
           &
           &
           \\
        \hline
        SegFormer B0 &
          56000 &
          94.3 &
          \textbf{94.4} &
          85.6 &
          \textbf{97.7} &
          \textbf{93.5} &
          94.6 &
          80.8 &
          92.5 &
          86.7 &
          \textbf{68.5} &
          92.7 \\
        SegFormer B1 &
          37000 &
          \textbf{94.5} &
          92.3 &
          \textbf{86.9} &
          97.3 &
          92.3 &
          \textbf{95.2} &
          \textbf{81.0} &
          \textbf{92.7} &
          \textbf{86.9} &
          65.6 &
          \textbf{92.9} \\
        \hline
    \end{tabular*}
\end{table*}

\subsection{Method comparison}
\label{subsec:methods-comparison}

Table~\ref{tbl:result_garden} shows the segmentation metrics of DT-SegNet against the state-of-the-art machine-learning methods. It can be clearly observed that the proposed DT-SegNet achieves the highest scores among all methods. Compared to the best model among the two software packages, Weka running \ac{FRF}, a significant advantage of DT-SegNet in terms of accuracy (4.2\%), precision (6.2\%), recall (23.0\%) and F1-score (18.2\%) can be observed. Furthermore, the proposed DT-SegNet exhibits a lower standard deviation on all metrics, showing substantial robustness. Compared to the best \ac{CNN} model, SegFormer B0, a 2.3\% improvement in the F1 score can be observed. The standard deviation of the proposed DT-SegNet is also lower. It is worth noticing that although \ac{CNN} models have achieved outstanding accuracy, they are weak in recall, which means more precipitates are missing. Low recall and high standard deviation may suggest that these methods lack the required robustness to handle the variety of different \ac{EM} images.

\begin{table*}[htb]
    \small
    \caption{\ Pixel-wise segmentation performance on our dataset is shown. All results are generated using their best workflows. Pixel classification metrics are used to make comparisons between multiple methods}
    \label{tbl:result_garden}
    \begin{tabular*}{\textwidth}{@{\extracolsep{\fill}}lllllll}
        \hline
        Method & Accuracy & Precision & Recall & \acs{IoU} & \acs{SSIM} & F1 \\
        \hline
        ilastik (\acs{LDA}~\cite{hastieElementsStatisticalLearning2009}) & $86.8 \pm 5.5$ & $40.0 \pm 17.5$ & $82.7 \pm 7.4$ & $35.9 \pm 13.8$ & $65.7 \pm 18.9$ & $51.6 \pm 16.2$ \\
ilastik (\acs{RF}~\cite{breimanRandomForests2001}) & $93.9 \pm 2.9$ & $63.7 \pm 19.8$ & $72.7 \pm 20.3$ & $49.5 \pm 14.5$ & $82.4 \pm 4.9$ & $65.2 \pm 13.3$ \\
ilastik (\acs{SVC}~\cite{plattProbabilisticOutputsSupport1999}) & $75.5 \pm 36.1$ & $49.4 \pm 34.0$ & $56.2 \pm 40.6$ & $23.5 \pm 18.4$ & $68.8 \pm 33.8$ & $35.3 \pm 23.1$ \\
        \midrule
        Weka (\acs{FRF}~\cite{breimanRandomForests2001}) & $93.5 \pm 6.3$ & $76.1 \pm 24.2$ & $69.6 \pm 17.0$ & $51.6 \pm 9.1$ & $82.7 \pm 10.3$ & $68.0 \pm 8.4$ \\
Weka (\acs{MLP}~\cite{kubatNeuralNetworksComprehensive1999}) & $92.0 \pm 6.1$ & $58.7 \pm 25.1$ & $79.5 \pm 11.8$ & $48.6 \pm 17.5$ & $70.9 \pm 17.1$ & $64.0 \pm 15.4$ \\
        \midrule
        U-Net~\cite{ronnebergerUNetConvolutionalNetworks2015} & $96.4 \pm 4.2$ & $80.5 \pm 14.9$ & $73.8 \pm 40.7$ & $63.9 \pm 35.5$ & $92.0 \pm 3.5$ & $71.3 \pm 38.7$ \\
        UNet 3+~\cite{huangUNetFullScaleConnected2020} & $96.3 \pm 4.1$ & $87.6 \pm 9.1$ & $70.4 \pm 38.9$ & $61.9 \pm 34.5$ & $91.5 \pm 3.5$ & $70.0 \pm 38.2$ \\
        DeepLabV3+~\cite{chenEncoderDecoderAtrousSeparable2018} & $97.6 \pm 1.2$ & $85.6 \pm 9.0$ & $87.1 \pm 6.1$ & $75.5 \pm 5.2$ & $92.4 \pm 1.8$ & $85.9 \pm 3.4$ \\
        \midrule
        SegFormer B0~\cite{xieSegFormerSimpleEfficient2021} & $98.1 \pm 0.6$ & $87.6 \pm 10.4$ & $88.9 \pm 6.2$ & $78.3 \pm 6.8$ & $93.3 \pm 1.1$ & $87.7 \pm 4.3$\\
        SegFormer B1~\cite{xieSegFormerSimpleEfficient2021} & $97.7 \pm 1.0$ & $84.1 \pm 12.0$ & $91.9 \pm 7.5$ & $77.1 \pm 5.3$ & $92.4 \pm 1.9$ & $87.0 \pm 3.4$\\
        \midrule
        \textbf{DT-SegNet} & $98.3 \pm 0.8$ & $87.8 \pm 8.2$ & $92.8 \pm 3.4$ & $81.9 \pm 5.6$ & $94.0 \pm 1.4$ & $90.0 \pm 3.3$ \\
        \hline
    \end{tabular*}
\end{table*}

It is also worth mentioning that statistic-based models like \ac{LDA} can detect most precipitates, resulting in high accuracy and recall. However, this approach induces more false-positive detections, leading to low precision and F1-score. Classical machine-learning-based models such as \ac{RF}, \ac{FRF}, and \ac{MLP}, however, have higher \ac{IoU} and accuracy but miss more precipitates.

To reduce the human bias in the manual dataset split process, as well as make the performance of the proposed DT-SegNet convincing, K-Fold cross-validation with five folds is performed. As shown in Table~\ref{tbl:k-fold}, the proposed model performs consistently on different dataset splits. In split 2, the test case has a completely different distribution from the training set, resulting in a slightly lower performance than other splits. Split 5, however, has a balanced distribution in two datasets, resulting in higher performance than other splits.

\begin{table*}[htb]
    \small
    \caption{\ Pixel-wise segmentation performance of proposed DT-SegNet in K-Fold cross-validation}
    \label{tbl:k-fold}
    \begin{tabular*}{\textwidth}{@{\extracolsep{\fill}}lllllll}
    \hline
    Split & Accuracy & Precision & Recall & \acs{IoU} & \acs{SSIM} & F1 \\
    \hline
    1 & $97.1 \pm 2.3$ & $90.5 \pm 5.6$ & $84.0 \pm 13.7$ & $76.3 \pm 9.4$ & $89.8 \pm 8.6$ & $86.3 \pm 6.4$ \\
    2 & $96.5 \pm 2.0$ & $79.1 \pm 10.7$ & $91.5 \pm 1.8$ & $73.6 \pm 9.1$ & $88.8 \pm 7.2$ & $84.5 \pm 6.0$ \\
    3 & $97.8 \pm 0.6$ & $86.8 \pm 8.1$ & $84.0 \pm 11.4$ & $73.4 \pm 5.4$ & $93.4 \pm 1.2$ & $84.6 \pm 3.5$ \\
    4 & $97.6 \pm 0.9$ & $83.9 \pm 8.1$ & $87.2 \pm 4.5$ & $75.0 \pm 8.9$ & $91.3 \pm 4.7$ & $85.5 \pm 5.7$ \\
    5 & $97.9 \pm 2.3$ & $90.7 \pm 4.6$ & $84.5 \pm 11.7$ & $77.9 \pm 11.5$ & $93.5 \pm 5.7$ & $87.2 \pm 7.4$ \\
    \midrule
    Avg & $97.4 \pm 1.7$ & $86.0 \pm 8.4$ & $86.3 \pm 9.3$ & $75.1 \pm 8.3$ & $91.3 \pm 5.8$ & $85.5 \pm 5.4$ \\
    \hline
    \end{tabular*}
\end{table*}

\subsection{Visual inspection}
\label{subsec:visualisation}

\begin{figure*}[p]
    \centering
    \includegraphics[width=0.9\textwidth]{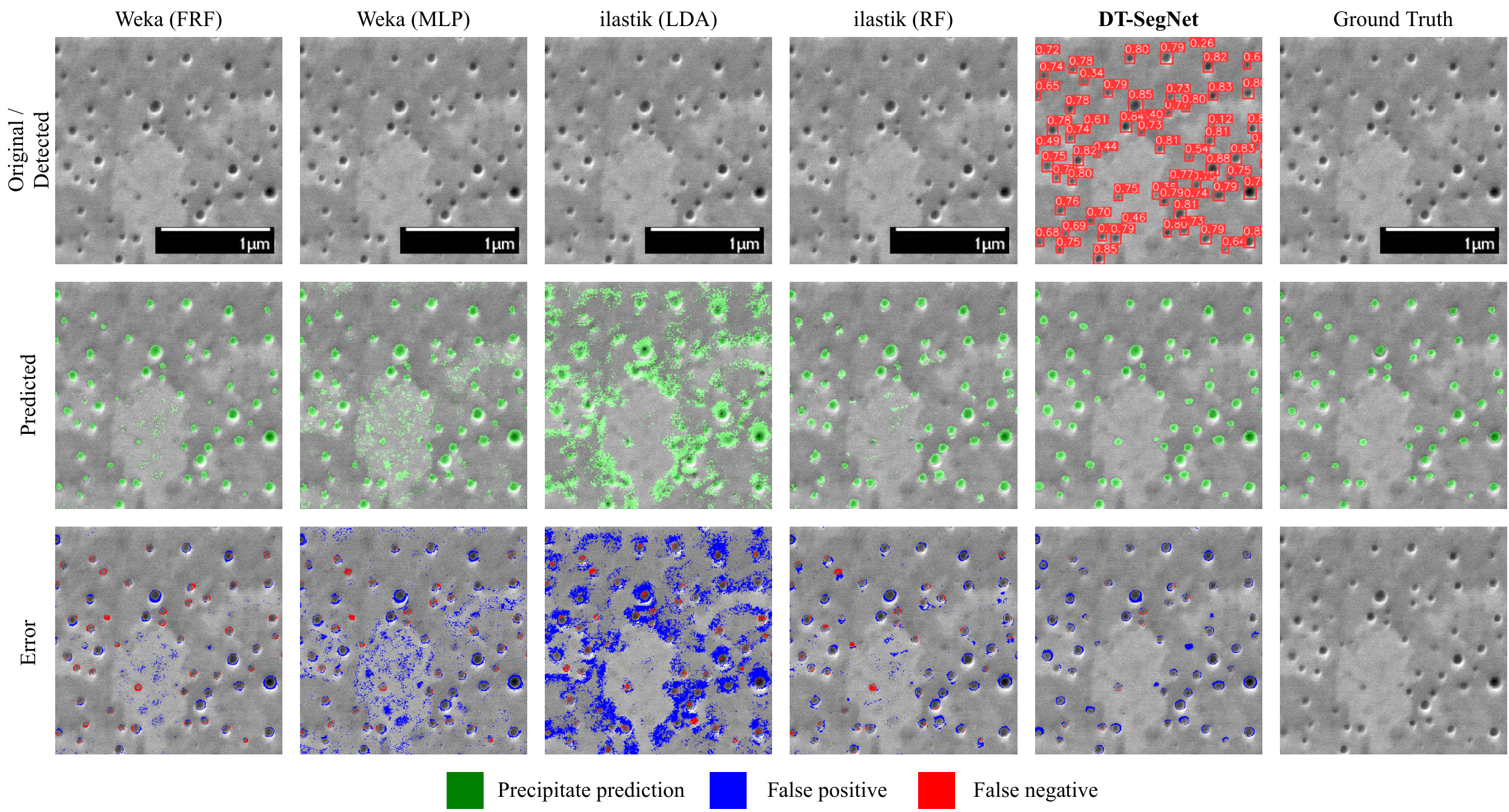}
    \caption{Visualisation of segmentation results on 5-5 produced by four competing methods and our methods, along with the ground truth annotation.}
    \label{fgr:result-14}
\end{figure*}

\begin{figure*}[p]
    \centering
    \includegraphics[width=0.9\textwidth]{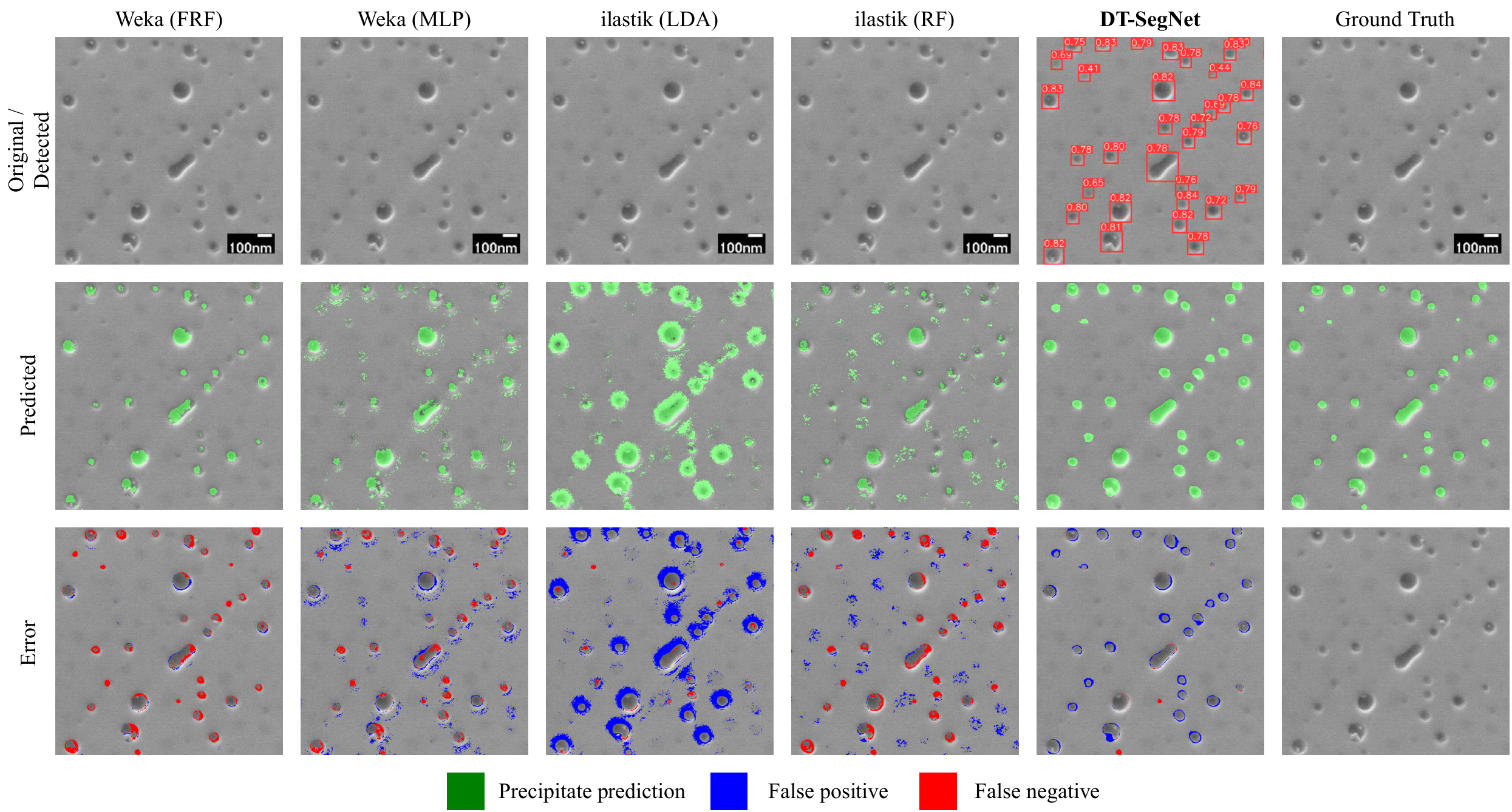}
    \caption{Visualisation of segmentation results on 5-5-10 produced by four competing methods and our methods, along with the ground truth annotation.}
    \label{fgr:result-1}
\end{figure*}

\begin{figure*}[p]
    \centering
    \includegraphics[width=0.9\textwidth]{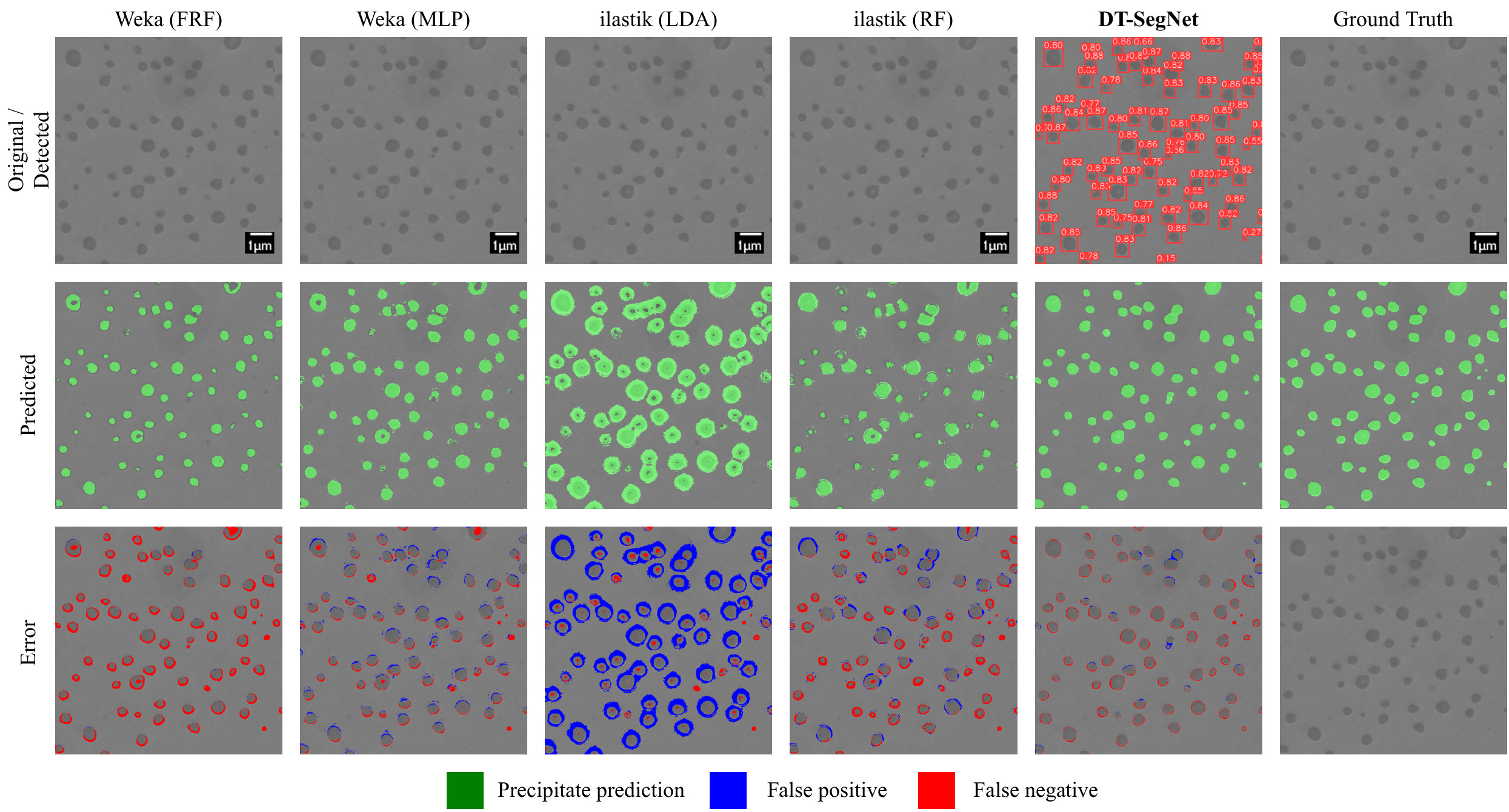}
    \caption{Visualisation of segmentation results on 10-10-20-4h produced by four competing methods and our methods, along with the ground truth annotation.}
    \label{fgr:result-5}
\end{figure*}

\begin{figure*}[p]
    \centering
    \includegraphics[width=0.9\textwidth]{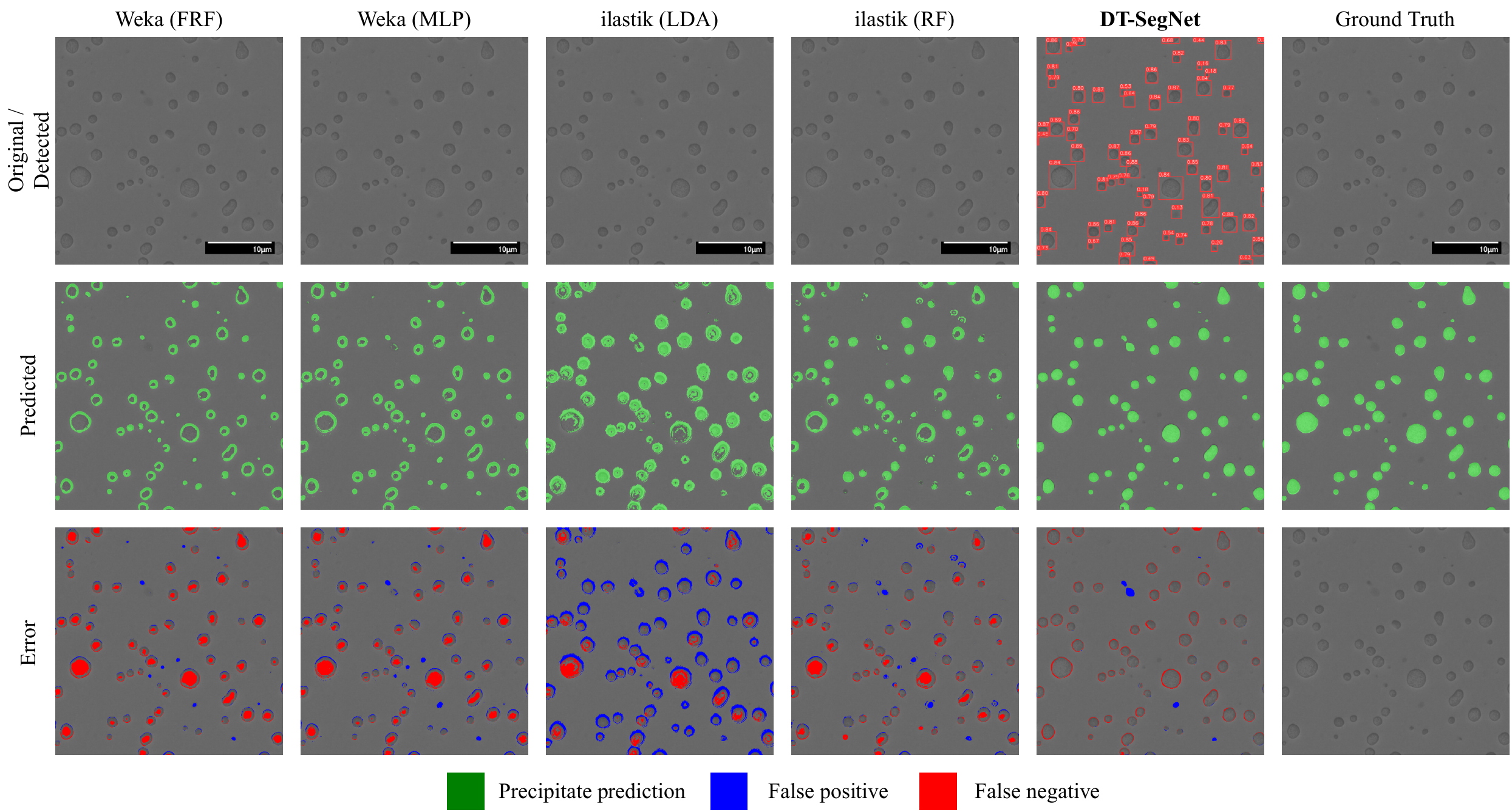}
    \caption{Visualisation of segmentation results on 10-10-20-100h produced by four competing methods and our methods, along with the ground truth annotation.}
    \label{fgr:result-20}
\end{figure*}

Segmentation quality can be most intuitively assessed by visualisation, as seen in Fig.~\ref{fgr:result-14},~\ref{fgr:result-1},~\ref{fgr:result-5}, and~\ref{fgr:result-20} of four \ac{SEM} images selected from the test set with outputs from different models. Conditions of the selected images are included in the training dataset.

The original input is shown in the first row, along with the ground truth annotation placed at the right of the first row. The output of the detection stage of DT-SegNet is also shown in the first row. The second row shows the models' predicted output; in this context, green represents the mask of the predicted precipitate. The background pixels are left as they are. For DT-SegNet, the best confidence threshold based on the performance of the validation set is used, while the other methods use their default confidence threshold. Perfect segmentation covers all the noticeable precipitates with the best-fitting shape. In the third row, a colourised illustration of taxonomy for segmented pixels is presented: false positive and negative predictions are marked in red. The fourth row shows the predicted precipitate area as a percentage of the original image. In the fifth row, the prediction error is given as a proportion of the input image.

Figure~\ref{fgr:result-14} shows a case with tremendous blurring and background noises frequently encountered in \ac{SEM} observations. Most methods except \ac{LDA} successfully detect all precipitates and segment them in good shape, with an error rate lower than 9\%. However, the other three baseline models have many false-positive predictions on the white background. It is worth mentioning that there is a spurious precipitate that most of the methods failed to ignore. The false-positive detection may be attributed to its darkness, which shows the real-world experiments' complexity. As a result, they are higher in error ratio compared with DT-SegNet. Although DT-SegNet detects a few background noises as precipitate, most are detected in low confidence and then filtered at the detection stage. Consequently, the segmentation stage only receives the \ac{ROI} as input, making the model more robust to the uncertain background.

Figure~\ref{fgr:result-1} is a common case of a \ac{SESEM} image showing nano-scale precipitates. The contrast inside precipitates is different from the contrast of the matrix. Due to the polishing, precipitates are polished slightly more than the matrix. It causes different heights on the precipitate area, which were clearly resolved using \ac{SE} imaging. Apart from the precipitates exposed on the surface, weak blurry contrast from some embedded precipitates is observed, which are excluded in the observation. It can be seen in the original images that the precipitates have white edges, which can be a helpful feature for models. Decision-tree-based algorithms like \ac{FRF} and \ac{RF} can detect most precipitates correctly and are the closest to the ground truth value, with errors near the edge. The error may be attributed to its lack of generalisation of objects in an irregular shape. \ac{LDA} fails to differentiate the edges of precipitates, so the detected area tends to be considerably larger than the ground truth. The \ac{MLP} produces a more robust result, but due to its small model size, the model has difficulties distinguishing the background noise from precipitates. DT-SegNet has perfect detection results on the input image (lowest error ratio), showing the model is robust to the background noises. However, it is still challenging for the model to fully detect small-scale precipitates, and the segmentation task of abnormal precipitates may still be inaccurate.

Figure~\ref{fgr:result-5} shows a case of the \ac{SESEM} image with nano-scale precipitates. In this figure, precipitates are larger than those in Fig.~\ref{fgr:result-1}, and the contrast is different. The edge is apparent, but some light points exist in these large precipitates. In this scenario, all models can better detect the precipitate area. However, both Weka- and ilastik-based methods fail to segment the exotic contrast in some precipitates due to the lack of robustness, which will affect the area measurement. The unstable interactive labelling mechanism of Weka and ilastik can cause this inability. On the other hand, DT-SegNet shows a substantially more accurate segmentation, achieving the lowest error rate of 2.28\%.

Figure~\ref{fgr:result-20} shows a case of \ac{SESEM} images with micro-scale precipitates. Despite the evident edges of precipitates, the contrast inside these precipitates is similar to the matrix. In this case, all four baseline models manage to detect the edges but show poor segmentation results on the textures inside the precipitates, showing error rates higher than 3.5\%. Since the segmentation network in DT-SegNet can capture most of the features, textures are well taken into account in the segmentation model, resulting in an outstanding performance of a 1.53\% error rate.

The online computational time of DT-SegNet averaged on the test dataset is shown in Table~\ref{tbl:effort}. The manual segmentation time is estimated for \ac{EM} images with 100 to 200 objects. It can be clearly seen that the proposed DT-SegNet can considerably improve the efficiency of precipitate segmentation compared to a manual process.

\begin{table}[htb]
    \small
    \caption{\ Process time of the proposed process and brute force manual. The manual segmentation time is estimated for \ac{EM} images with 100 to 200 objects}
    \label{tbl:effort}
    \begin{tabular*}{0.48\textwidth}{@{\extracolsep{\fill}}llll}
        \hline
        \multicolumn{3}{c}{DT-SegNet} & Manual \\
        \cmidrule(lr){1-3}\cmidrule(lr){4-4}
        Detection & Segmentation & Total & Total \\
        \hline
        0.0214s & 1.8148s & 2.3718s & $\approx 30$ min\\
        \hline
    \end{tabular*}
\end{table}

Overall, the proposed DT-SegNet considerably outperforms all Weka- and ilastik-based state-of-the-art approaches for multi-scale precipitate detection and area measurement from \ac{SEM} images along with various background contrast.

\subsection{Microstructural analysis of Cr-superalloys}
\label{subsec:microstructure-analysis}

Table~\ref{tbl:four-result} presents the results of the area fraction and average radius of precipitates measured manually (ground truth) and using the proposed DT-SegNet method. The two measurements are in good agreement, as discussed in the previous section. Here it is assumed that the volume fraction of precipitates equals the area fraction. It is worth noting that the two 10-10-20 alloys have higher precipitate volume fraction than the 5-5 and 5-5-10. The volume fraction is a key factor in pursuing high strength in these superalloys, as the precipitate strengthening, including ordering, coherency, modulus, and Orowan strengthening, increases with volume fraction~\cite{schwarzDynamicSimulationSolution1978,reppichAttractiveParticleDislocationInteraction1998,brownWorkhardeningCoppersilica1971,nembachHardeningCoherentPrecipitates1984,kocksTheoryObstacleControlledYield1977}. Meanwhile, 10-10-20-4h has smaller precipitates than 10-10-20-100h due to the precipitate coarsening at $1200^{\circ}\text{C}$.

\begin{table*}[htb]
    \small
    \caption{\ Area fraction and average radius of precipitates by manual measurements and by DT-SegNet}
    \label{tbl:four-result}
        \begin{tabular*}{\textwidth}{@{\extracolsep{\fill}}lllll}
        \hline
        \multirow{2}{*}{Image} & \multicolumn{2}{c}{Area Fraction (\%)} & \multicolumn{2}{c}{Radius (${nm})$} \\\cmidrule(lr){2-3}\cmidrule(lr){4-5}
            & DT-SegNet & Ground Truth & DT-SegNet & Ground Truth \\
        \hline
        5-5        & $8.92$ & $6.47$ & $37.30 \pm 9.85$ & $32.66 \pm 6.63$    \\
        5-5-10      & $5.33$ & $3.72$ & $34.86 \pm 11.54$ & $29.33 \pm 11.31$   \\
        10-10-20-4h  & $8.99$ & $10.00$ & $210.09 \pm 63.89$ & $229.86 \pm 63.60$  \\
        10-10-20-100h & $10.03$ & $11.66$ & $695.61 \pm 267.50$ & $752.23 \pm 287.76$ \\
        \hline
        \end{tabular*}
\end{table*}

\begin{figure*}[htb]
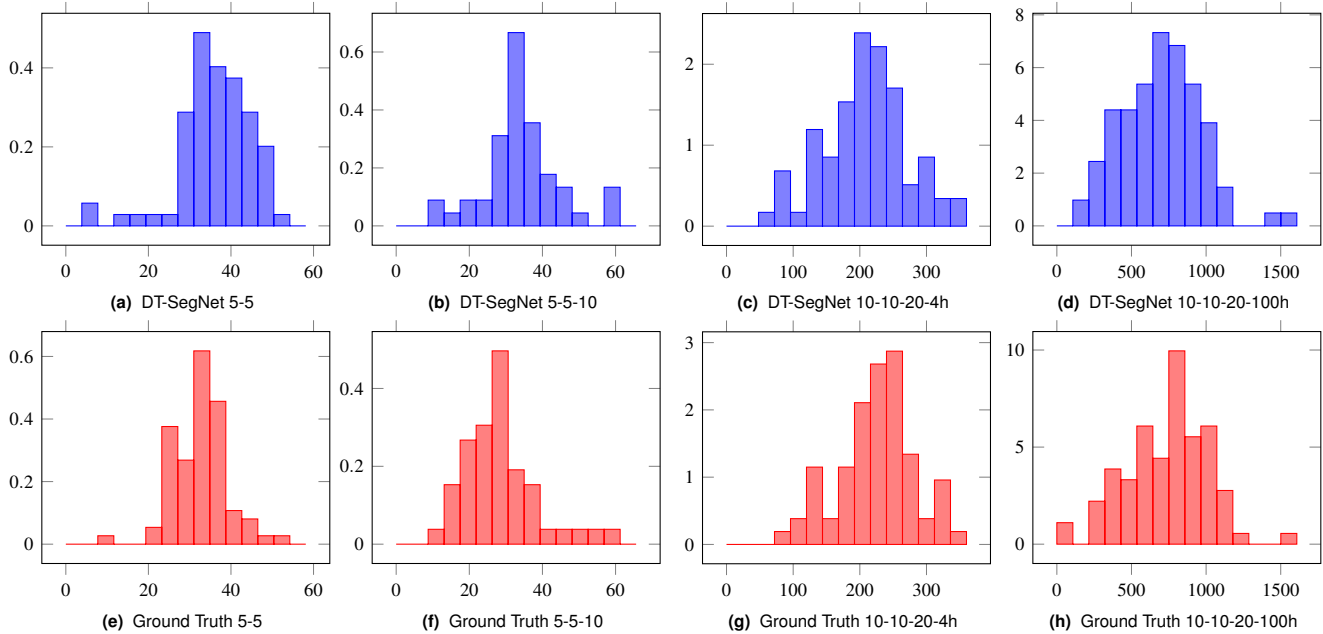

    \centering

    \subcaptionbox{\label{fgr:gardenhist-ours-14} DT-SegNet 5-5}
    {\rhoSqHInner{gardenhist_ours_14.txt}{blue}\vspace{-.5\baselineskip}\hspace{\baselineskip}}
    \subcaptionbox{\label{fgr:gardenhist-ours-1} DT-SegNet 5-5-10}
    {\rhoSqHInner{gardenhist_ours_1.txt}{blue}\vspace{-.5\baselineskip}\hspace{\baselineskip}}
    \subcaptionbox{\label{fgr:gardenhist-ours-5} DT-SegNet 10-10-20-4h}
    {\rhoSqHInner{gardenhist_ours_5.txt}{blue}\vspace{-.5\baselineskip}\hspace{\baselineskip}}
    \subcaptionbox{\label{fgr:gardenhist-ours-20} DT-SegNet 10-10-20-100h}
    {\rhoSqHInner{gardenhist_ours_20.txt}{blue}\vspace{-.5\baselineskip}\hspace{\baselineskip}}

    \subcaptionbox{\label{fgr:gardenhist-gt-14} Ground Truth 5-5}
    {\rhoSqHInner{gardenhist_gt_14.txt}{red}\vspace{-.5\baselineskip}\hspace{\baselineskip}}
    \subcaptionbox{\label{fgr:gardenhist-gt-1} Ground Truth 5-5-10}
    {\rhoSqHInner{gardenhist_gt_1.txt}{red}\vspace{-.5\baselineskip}\hspace{\baselineskip}}
    \subcaptionbox{\label{fgr:gardenhist-gt-5} Ground Truth 10-10-20-4h}
    {\rhoSqHInner{gardenhist_gt_5.txt}{red}\vspace{-.5\baselineskip}\hspace{\baselineskip}}
    \subcaptionbox{\label{fgr:gardenhist-gt-20} Ground Truth 10-10-20-100h}
    {\rhoSqHInner{gardenhist_gt_20.txt}{red}\vspace{-.5\baselineskip}\hspace{\baselineskip}}

    \vspace{-.5\baselineskip}
    \caption{The $ \rho^2h(\rho) $ particle size distribution of the four studied materials.}
    \label{fgr:gardenhist}
\end{figure*}

Furthermore, analogous to some ferritic superalloys (Fe--NiAl systems) with a similar structure as the Cr--NiAl alloys\cite{cayetano-castroOstwaldRipeningProcess2015,calderonCoarseningKineticsCoherent1984}, it is assumed that the precipitates in these Cr-superalloys underwent diffusion-controlled coarsening during the used heat treatment condition. The particle size distribution (PSD) is plotted in Fig.~\ref{fgr:gardenhist}. The co-ordinates are the probability density $\rho^{2}h(\rho)$ which is calculated as:

\begin{equation}
    \rho^{2}h(\rho)=\frac{N_{r,r+\Delta r}}{\Sigma N_{r,r+\Delta r}}\frac{\Bar{r}}{\Delta r}
\end{equation}

where $N_{r,r+\Delta r}$ is the number of precipitate in each interval, $\Bar{r}$ is the average radius of precipitates and $\Delta r$ is the bin size of the distribution analysis. The two 10-10-20 alloys show a larger average radius suggesting a higher coarsening rate in 10-10-20 alloys than the 5-5 and 5-5-10. Along with the ageing, 10-10-20-100h shows a broader distribution than 10-10-20-4h as a result of precipitate coarsening, as observed in other A2-B2 systems like Fe--NiAl alloys~\cite{cayetano-castroOstwaldRipeningProcess2015,calderonCoarseningKineticsCoherent1984,sunNanoSizedPrecipitateStability2015,dorantes-rosalesPrecipitationProcessFeNiAlBased2015}.

\balance

It is also worth noting that the ground truth only provides reference values for comparison among different segmentation methods and could be user-dependent. The measured values of the precipitate area and radius from \ac{SEM} images by all methods are systematically smaller than their absolute values as the area of precipitates exposed to the surface is systematically smaller or equal to the largest cross-section of the precipitate sphere. Geometric correction for radius could be used to correct this bias~\cite{sunNanoSizedPrecipitateStability2015, baikEffectHafniumMicroaddition2018}. Other frequently used imaging techniques, such as \ac{TEM} could also provide similar measurements with different biases. The application of the current detection and segmentation method would also be of great interest for precipitate size analysis by \ac{TEM}.

\section{Conclusion}
\label{sec:conclusion}

Efficient and accurate object detection, as well as segmentation, are important for \ac{EM} image analysis when developing novel materials, and are critical to handle the large datasets associated with high-throughput combinatorial discovery methods. Traditional approaches consist of filtering \ac{EM} images with a contrast threshold. However, the robustness of such a method can be challenged under different experimental conditions/noises, and often requires laborious manual adjustments.

In this work, a two-stage end-to-end deep learning scheme, DT-SegNet, using state-of-the-art deep learning frameworks is proposed, namely YOLOv5 for object detection and SegFormer for segmentation.

The model has been applied for precipitate pixel segmentation in novel Cr-superalloys, which comprise a two-phase microstructure of an A2 Cr matrix with B2 NiAl spherical precipitates, developed for high-temperature applications such as advanced Concentrated Solar Power. The precipitate size and volume fraction are important factors controlling the mechanical properties in the superalloys. Extensive numerical experiments have shown the strength of DT-SegNet compared to the state-of-the-art tools Weka and ilastik in a number of different metrics, including accuracy, standard deviation, recall, F1-score and \ac{SSIM}. Furthermore, DT-SegNet is only trained using 15 images in this application. Thus, the proposed approach can be easily applied/transferred to other materials using a small amount of data for fine-tuning. The DT-SegNet method is applied in the development of new Cr(Fe)--NiAl alloys for high-temperature applications. Area fraction, average radius and size distribution of precipitates were measured in different alloys where the precipitate size varies from nano-scale to micro-scale. In this multi-scale measurement, results from the DT-SegNet method show a good agreement with the manual measurement. 

Future efforts can be considered to train the neural networks of detection and segmentation jointly so that the model fine-tuning for new materials can be further simplified. The tuned model will be further used for the determination of the precipitate coarsening rate of Cr-superalloys by measuring the precipitate size as a function of the ageing time for a given temperature. 
The current training dataset can be expanded to datasets including not only Cr-superalloys but also other advanced alloy systems, accelerating alloy development and microstructure examination. Furthermore, such low user intervention models are critical tools to enable the analysis of large datasets from high-throughput combinatorial metallurgy.

\section*{Code and data availability}

The computational part of this study is performed using Python language. The code and the \ac{EM} data used in this study are available at: \url{https://doi.org/10.5281/zenodo.7510032}.

\clearpage
\section*{Acronyms}
\label{sec:acronyms}

\begin{acronym}[CALPHAD]
    \acro{Al}{Aluminium}
    \acro{ASPP}{Atrous Spatial Pyramid Pooling}
    \acro{AP}{Average Precision}
    \acro{bcc}{body-centred-cubic}
    \acro{fcc}{face-centred-cubic}
    \acro{CALPHAD}{CALculation of PHAse Diagram}
    \acro{CNN}{Convolutional Neural Network}
    \acro{COCO}{Common Objects in Context}
    \acro{Cr}{Chromium}
    \acro{CSP}{Cross Stage Partial}
    \acro{DNN}{Deep Neural Network}
    \acro{EM}{Electron Microscopy}
    \acro{FCNN}{Fully Convolutional Neural Network}
    \acro{FFN}{Feed Forward Network}
    \acro{FRF}{Fast Random Forest}
    \acro{ICME}{Integrated Computational Materials Engineering}
    \acro{IoU}{Intersection over Union}
    \acro{LDA}{Linear Discriminant Analysis}
    \acro{mAP}{mean Average Precision}
    \acro{mIoU}{mean Intersection over Union}
    \acro{MLP}{Multi-Layer Perceptron}
    \acro{MGI}{Materials Genome Initiative}
    \acro{Ni}{Nickel}
    \acro{Fe}{Iron}
    \acro{NiAl}{Nickel--Aluminide}
    \acro{OPS}{Oxide Polishing Suspensions}
    \acro{PANet}{Path Aggregation Network}
    \acro{PRC}{Precision-Recall Curve}
    \acro{RF}{Random Forest}
    \acro{ROI}{Region of Interest}
    \acro{SE}{Secondary Electron}
    \acro{SEM}{Scanning Electron Microscope}
    \acro{SESEM}{Secondary Electron Scanning Electron Microscope}
    \acro{SGD}{Stochastic Gradient Descent}
    \acro{SPP}{Spatial Pyramid Pooling}
    \acro{SPPF}{Spatial Pyramid Pooling Fast}
    \acro{SSIM}{Structural Similarity Index}
    \acro{SVM}{Support Vector Machine}
    \acro{SVC}{Support Vector Machines C-Support}
    \acro{TEM}{Transmission Electron Microscopy}
    \acro{ViT}{Vision Transformer}
    \acro{YOLO}{You Only Look Once}
\end{acronym}

\section*{Conflicts of interest}
The authors have no conflicts of interest to disclose.

\section*{Acknowledgements}
This project has received funding from the European Union's Horizon 2020 research and innovation programme under grant agreement No 958418 ``COMPASsCO2'' (https://www.compassco2.eu). The authors thank the Centre for Electron Microscopy (University of Birmingham) for their support and assistance in this work. This work is partially supported by the EP/T000414/1 PREdictive Modelling with Quantification of UncERtainty for MultiphasE Systems (PREMIERE).

\bibliography{references}
\bibliographystyle{rsc}

\end{document}